\documentclass[5pt]{article}

\usepackage[letterpaper]{geometry}
\usepackage{spconf,amsmath,epsfig}
\usepackage{enumitem}
\usepackage{amssymb}
\usepackage{latexsym}
\usepackage{subfigure}
\usepackage{times}
\usepackage{url}
\usepackage{graphicx}
\usepackage{scalerel}
\usepackage{amsmath}
\usepackage{amssymb}
\usepackage{caption}
\usepackage{array}
\usepackage{enumitem}
\usepackage{algorithm}
\usepackage{algpseudocode}
\usepackage{booktabs}
\usepackage{multirow}
\usepackage{rotating}
\usepackage{algpseudocode}
\usepackage{tabularx}
\usepackage{dblfloatfix}
\usepackage{optidef}
\usepackage{bm}
\usepackage{cleveref}

\DeclareMathOperator*{\argmax}{\arg\,\max}

\newcommand{\Cbf}{\mathbf{C}}

\newcommand{\Cbfhat}{\hat{\mathbf{C}}}

\newcommand{\Dbf}{\mathbf{D}}

\newcommand{\Ebf}{\mathbf{E}}

\newcommand{\Ibf}{\mathbf{I}}

\newcommand{\mbf}{\mathbf{m}}

\newcommand{\Rbb}{\mathbb{R}}

\newcommand{\Ubf}{\mathbf{U}}
\newcommand{\Ubfhat}{\hat{\mathbf{U}}}

\newcommand{\xbf}{\mathbf{x}}
\newcommand{\Xbf}{\mathbf{X}}

\newcommand{\Ybf}{\mathbf{Y}}

\newcommand{\zbf}{\mathbf{z}}

\newcommand{\Lambdabf}{\mathbf{\Lambda}}

\newcommand{\onebf}{\textbf{1}}
\newcommand{\onebft}{\textbf{1}^{\!\top}}
\newcommand{\msh}{\!\!\:}

\usepackage{url}
\usepackage{xcolor}
\definecolor{newcolor}{rgb}{.8,.349,.1}
\definecolor{L1TColor}{rgb}{0,0,0}
\definecolor{StruckColor}{rgb}{1,0,1}
\definecolor{IVTColor}{rgb}{0,1,1}
\definecolor{MTTColor}{rgb}{0.5,0.5,0.5}
\definecolor{MILColor}{RGB}{136,0,21}

\definecolor{VTDColor}{RGB}{255,127,39}
\definecolor{FragColor}{RGB}{0,162,232}
\definecolor{ASLAColor}{RGB}{163,73,164}
\definecolor{KCFColor}{rgb}{1,0,0}
\definecolor{MEEMColor}{rgb}{0,1,0}
\definecolor{RSST_HOGColor}{rgb}{1,0,0}
\definecolor{SGLST_ColorColor}{rgb}{0,1,0}
\definecolor{SGLST_HOGColor}{rgb}{0,0,1}

\usepackage{fancyhdr}
\begin{document}\sloppy

\def\x{{\mathbf x}}
\def\L{{\cal L}}

\title{Structured Group Local Sparse Tracker}
%
\name{Mohammadreza Javanmardi, Xiaojun Qi}
%
\address{Department of Computer Science, Utah State University, Logan, UT 84322-4205, USA}
%
%

\maketitle

\begin{abstract}
Sparse representation is considered as a viable solution to visual tracking. In this paper, we propose a structured group local sparse tracker (SGLST), which exploits local patches inside target candidates in the particle filter framework.  Unlike the conventional local sparse trackers, the proposed optimization model in SGLST not only adopts local and spatial information of the target candidates but also attains the spatial layout structure among them by employing a group-sparsity regularization term. To solve the optimization model, we propose an efficient numerical algorithm consisting of two subproblems with the closed-form solutions. Both qualitative and quantitative evaluations on the benchmarks of challenging image sequences demonstrate the superior performance of the proposed tracker against several state-of-the-art trackers.
\end{abstract}

\maketitle

\section{Introduction}\label{sec:intro}

Visual tracking is the process of estimating states of a moving object in a dynamic frame sequence. It has been considered as one of the most paramount and challenging topics in computer vision with various applications in human motion analysis, surveillance, smart vehicles transportation, navigation, etc. Although numerous tracking methods \cite{yilmaz2006object,salti2012adaptive,pang2013finding,kristan2015visual,zhang2015structural,zhang2016defense} have been introduced in recent years, developing a robust algorithm that can handle different challenges such as occlusion, illumination variations, deformation, fast motion, camera motion, and background clutter still remains unsolved. 

Visual tracking algorithms can be roughly classified into discriminative and generative categories. Discriminative approaches cast the tracking problem as binary classification and formulate a decision boundary to separate the target from backgrounds. Representative discriminative approaches include the ensemble tracker \cite{avidan2007ensemble}, the online boosting \cite{grabner2006real,grabner2008semi}, the multiple instance learning \cite{babenko2009visual}, the PN-learning \cite{kalal2010pn}, and correlation filter-based trackers \cite{danelljan2014accurate,ma2015long,hu2017correlation}. In contrast, generative approaches adopt a model to represent the target and cast the tracking as a searching procedure to find the most similar region to the target model.  Representative generative tracking methods include eigen-tacking \cite{black1998eigentracking}, mean-shift  \cite{comaniciu2003kernel}, Frag-Track \cite{adam2006robust}, incremental learning \cite{ross2008incremental}, visual tracking decomposition \cite{kwon2010visual}, and adaptive color tracking \cite{roffo2016online}.

Sparse representation based trackers (sparse trackers) are considered as generative tracking methods since they sparsely express the target candidates using a few templates (bases). Generally, sparse representation has played a dominant role in computer vision applications such as face recognition \cite{chao2011locality}, image denoising and restoration \cite{li2012group}, image segmentation \cite{zohrizadeh2018image, zohrizadeh2016reliability}, image pansharpening \cite{Tang2017PansharpeningVL}, etc. Most sparse trackers utilize a convex optimization model to represent the global appearance of target candidates in the particle filter framework. As one of the pioneer work, Mei et al. \cite{mei2011robust} represent the global information of target candidates by a set of templates using $\ell_1$ minimization. Bao et al. \cite{bao2012real} present an accelerated proximal gradient descent method to increase the efficiency of solving $\ell_1$ minimization. In order to attain the relationship among target candidates, Zhang et al. \cite{zhang2012low} propose to jointly learn the global information of all target candidates. Later, Hong et al. \cite{hong2013tracking} cast tracking as a multi-task multi-view sparse learning problem in terms of the least square (LS). To handle the data possibly contaminated by outliers and noise, Mei et al. \cite{mei2015robust} use the least absolute deviation (LAD) in their optimization model. In general, these global sparse trackers achieve good performance. However, they model each target region as a single entity and may fail when targets undergo heavy occlusions in a frame sequence. 

Unlike global sparse trackers, local sparse trackers represent local patches inside target candidates together with local patches inside each template set. Liu et al. \cite{liu2013robust} introduce a local sparse tracker, which adopts the histogram of sparse coefficients and a sparse constrained regularized mean-shift algorithm, to robustly track the object. This method is based on a static local sparse dictionary and therefore fails in the cases when similar objects appear in the scene. Jia et al. \cite{jia2012visual} exploit both partial and spatial information of target candidates and represent them in a dynamic local sparse dictionary. More recently, Jia et al. \cite{jia2016visual} propose to extract coarse and fine local image patches inside each target candidate. Despite favorable performance, these local sparse trackers \cite{jia2012visual, jia2016visual} do not consider the spatial layout structure among local patches inside a target candidate. As a result, the sparse vectors of local patches exhibit a random pattern rather than a similar structure on the non-zero elements. 

To further improve the tracking performance, recent sparse trackers consider both global and local information of all target candidates in their optimization models. Zhang et al. \cite{zhang2015structural} represent local patches inside all target candidates along with the global information using $\ell_{1,2}$ norm regularization on the sparse representative matrix. They assume that the same local patches of all target candidates are similar. However, this assumption does not hold in practice due to outlier candidates and occlusion in tracking. To address this shortcoming, Zhang et al. \cite{zhang2018robust} take into account both factors to design an optimal target region searching method. These recent sparse trackers achieve improved performance. However, considering the relationship of all target candidates degrades the performance when drifting occurs. In addition, using $\ell_{1,2}$ norm regularization in the optimization model to integrate both local and global information of target candidates lessens the tracking accuracy in the cases of heavy occlusions.

In this paper, we propose a structured group local sparse tracker (SGLST), which exploits local patches inside a target candidate and represent them in a novel convex optimization model. The proposed optimization model not only adopts local and spatial information of the target candidates but also attains the spatial layout structure among them by employing a group-sparsity regularization term. The main contributions of the proposed work are summarized as follows:
\begin{itemize}
\item Proposing a local sparse tracker, which employs local and spatial information of a target candidate and attains the spatial structure among different local patches inside a target candidate.

\item Developing a convex optimization model, which introduces a group-sparsity regularization term to motivate the tracker to select the corresponding local patches of the same small subset of templates to represent the local patches of each target candidate. 

\item Designing a fast and parallel numerical algorithm based on the alternating direction method of multiplier (ADMM), which consists of two subproblems with the closed-form solutions. 
\end{itemize}

The remainder of this paper is organized as follows: Section \ref{sec:not} introduces the notations. Section \ref{sec:pro} presents the SGLST together with its novel convex optimization model solved by the proposed ADMM-based numerical solution. Section \ref{sec:exp} demonstrates the experimental results on 16 publicly challenging image sequences, the OTB50, and the OTB100 tracking benchmarks and compares the SGLST with several state-of-the-art trackers. Section \ref{sec:conc} draws the conclusions.
\section{Notations}\label{sec:not}
Throughout this paper, matrices, vectors, and scalers are denoted by boldface uppercase, boldface lowercase, and italic lowercase letters, respectively. For a given matrix $\Xbf$, $\Xbf_{i,j}$ denotes the element at the $i^{th}$ row and $j^{th}$ column, $\left\|\Xbf\right\|_{\mathrm{F}}$ indicates the Frobenious norm, $\left\|\Xbf\right\|_{p,q}$ is the $\ell_p$ norm of $\ell_q$ norm of the rows in $\Xbf$, and $\Xbf(:)$ is the vectorized form of $\Xbf$. For a given column vector $\xbf$, $\mathrm{diag}(\xbf)$ and $\xbf_i$ denote a diagonal matrix formed by the elements of $\xbf$ and the $i^{th}$ element of $\xbf$, respectively. Symbol $\mathrm{tr}(\cdot)$ stands for the trace operator, $\Xbf\otimes\Ybf$ is the Kronecker product on two matrices $\Xbf$ and $\Ybf$ of arbitrary sizes, $\onebf_{l}$  is a column vector of all ones with the dimension of $l$, and $\Ibf_{k}$ denotes a $k \times k$ identity matrix.
\section{Proposed Method} \label{sec:pro}
This section provides detailed information about the proposed structured group local sparse tracker (SGLST). Specifically, subsection \ref{subsec:Formulation} formulates a local sparse appearance model in SGLST and explains how this convex optimization model addresses the drawbacks of conventional local sparse trackers \cite{jia2012visual,jia2016visual}. Subsection \ref{subsec:NumericAlg} presents an efficient numerical algorithm to solve the convex optimization problem presented in subsection \ref{subsec:Formulation}.

\subsection{Structured Group Local Sparse Tracker (SGLST)} \label{subsec:Formulation}
The proposed SGLST utilizes both local and spatial information in the particle filter framework and employs a new optimization model, which addresses the drawback of conventional local sparse trackers by attaining the spatial layout structure among different local patches inside a target candidate.
\begin{figure*}[t]
\centering     
\subfigure[]{\label{fig:1a}\includegraphics[width=0.49\textwidth]{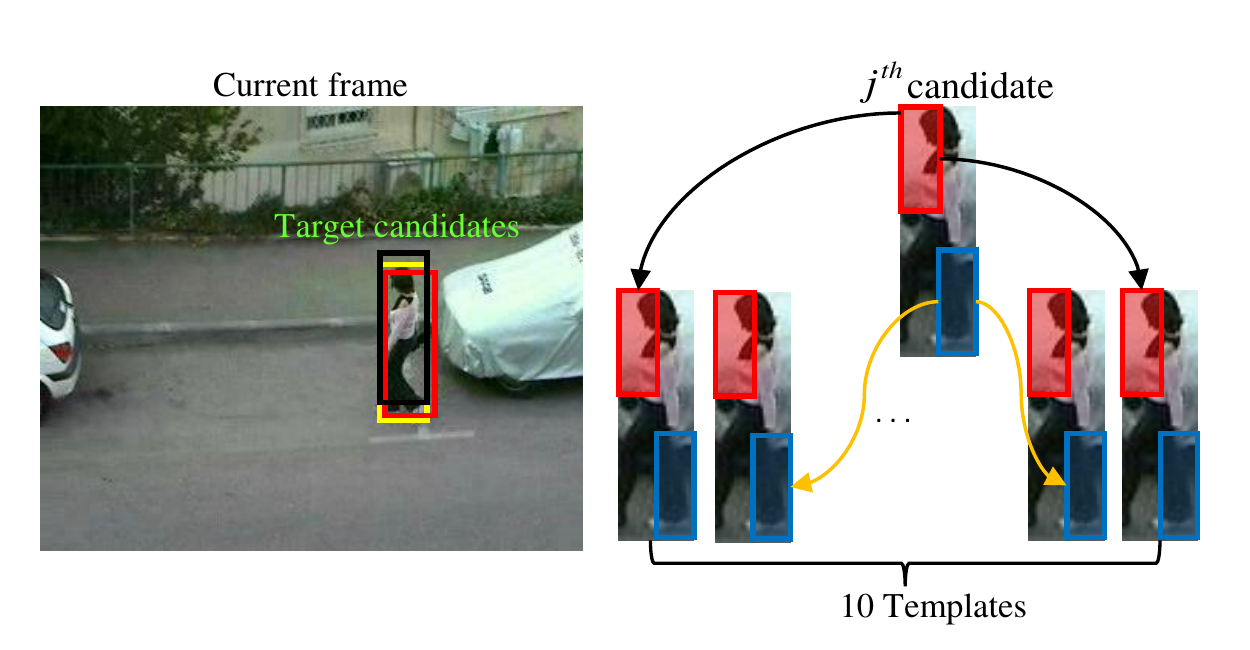}}
\subfigure[]{\label{fig:1b}\includegraphics[width=0.49\textwidth]{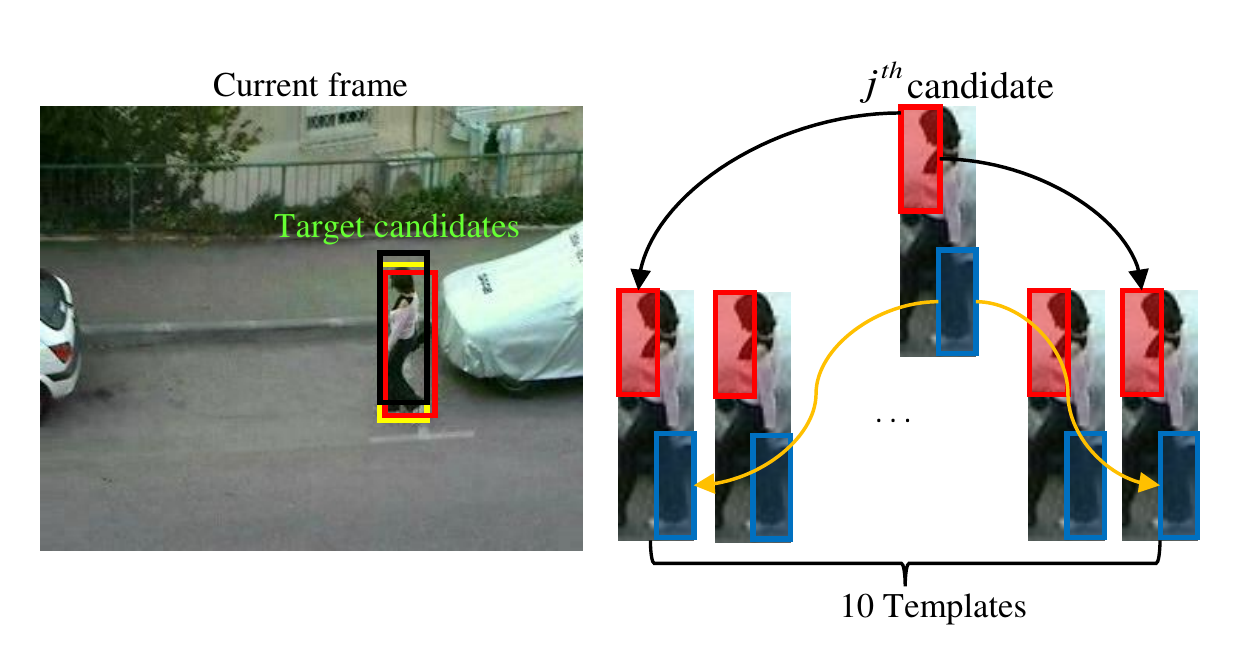}}
\caption{Illustration of the sparse representation of two sample local patches of the $j^{th}$ target candidate in: (a) Conventional local sparse trackers \cite{jia2012visual,jia2016visual}. One local patch of the $j^{th}$ target candidate, shown in the red bounding box, is represented by its corresponding patch in the first and the tenth templates, while another local patch of this candidate, shown in the blue bounding box, is represented by its corresponding patch in two different templates (e.g., the second and the ninth templates). (b) The proposed SGLST.  Both local patches of the $j^{th}$ target candidate, shown in red and blue bounding boxes, are represented by their corresponding patches in the same templates (e.g., the first and the tenth templates).}
\label{fig:fig1}
\end{figure*}

Conventional local sparse  trackers \cite{jia2012visual, jia2016visual} individually represent local patches without considering their spatial layout structure. For instance, local patches in \cite{jia2012visual} are separately represented by solving the Lasso problem. As a consequence, local patches inside the $j^{th}$ target candidate may be sparsely represented by the corresponding local patches inside \textit{different} dictionary templates, as illustrated in Figure \ref{fig:1a}, where two local patches of the $j^{th}$ target candidate, shown in the red and blue bounding boxes, may be represented by the corresponding local patches in different dictionary templates. 

In this paper, we propose a novel SGLST that adopts both local and spatial information of the target candidates for tracking. The proposed tracker employs a novel optimization model to solve the aforementioned issues associated with conventional local sparse trackers \cite{jia2012visual,jia2016visual}. Specifically, SGLST formulates an optimization problem to impose a structure on the achieved sparse vectors for different local patches inside each target candidate and attain the spatial layout structure among the local patches. To solve the proposed model, we develop an efficient numerical algorithm consisting of two subproblems with closed-form solutions by adopting the alternating direction method of multiplier (ADMM) within each target candidate in the optimization function. 
To maintain the spatial layout structure among local patches, we jointly represent all the local patches of a target candidate in a new convex optimization model. In other words, if the $r^{th}$ local patch of the $j^{th}$ target candidate is best represented by the $r^{th}$ local patch of the $q^{th}$ template, the $s^{th}$ local patch of the $j^{th}$ target candidate should also be best represented by the $s^{th}$ local patch of the $q^{th}$ template. As shown in Figure \ref{fig:1b}, we aim to represent both local patches of the $j^{th}$ target candidate, shown in the red and blue bounding boxes, by their corresponding patches in the \textit{same} dictionary templates (e.g., the first and the tenth templates). 

To do so, we first use $k$ target templates and extract $l$ overlapping $d$ dimensional local patches inside each template to construct the dictionary $\Dbf$. Such a representation generates the local dictionary matrix $\Dbf=[\Dbf_1,\dots,\Dbf_k]\in\Rbb^{d\times (l k)}$, where $\Dbf_i \in\Rbb^{d\times l}$. Then, we construct a matrix $\Xbf=[\Xbf_1,\dots,\Xbf_n]\in \Rbb^{d\times (l n)}$, which contains the local patches of all the target candidates, where $n$ is the number of particles. Next, we define the sparse matrix coefficients $\Cbf$ corresponding to the $j^{th}$ target candidate as $\Cbf \triangleq {\begin{bmatrix} \Cbf_{1} & \cdots & \Cbf_{k} \end{bmatrix}}^{\top} \in {\Rbb}^{{(lk)} \times {l}}$, where $\{\Cbf_{q}\}_{q=1}^{k}$ is a $l\times l$ matrix indicating the group sparse representation of $l$ local patches of the $j^{th}$ target candidate using $l$ local patches of the $q^{th}$ template. Finally, we formulate the following convex model:
\begin{mini!} 
{\substack{\Cbf \in{\Rbb}^{{(lk)} \times {l}}}}{{\left\|\Xbf_j\!-\!\Dbf\Cbf\right\|}_{\mathrm{F}}^2\!+\!\lambda {\left\|{\begin{bmatrix} {\Cbf_{1}(:)}  \ldots {\Cbf_{k}(:)} \end{bmatrix}}^{\msh\top}\! \right\|}_{1,\infty}} 
		{\label{eq:OurModel1}}{\label{eq:OurModels1}}
		\addConstraint {\Cbf\geq 0}\label{eq:OurModels12}
        \addConstraint {{\onebf}^{\top}_{lk}} \Cbf = {{\onebf}^{\top}_{l}}, \label{eq:OurModels11}
\end{mini!}
where the first term corresponds to the total cost of representing feature matrix $\Xbf_j$ using the dictionary matrix $\Dbf$ and the second term is a group-sparsity regularization term, which penalizes the objective function in proportion to the number of selected templates (dictionary words). Moreover, the group-sparsity regularization term imposes all the local patches to jointly select similar few templates by simultaneously establishing the $\left\|\cdot\right\|_{1,\infty}$ minimization on matrix $\Cbf$. The regularization parameter $\lambda>0$ balances the trade-off between the two terms. The constraint \eqref{eq:OurModels11} ensures that each local patch in $\Xbf_j$ is expressed by at least one selected local patch of the dictionary $\Dbf$ and the sum of a linear combination of coefficients is constrained. 
 
For each target candidate, we find the sparse matrix $\Cbf$ using the numerical algorithm presented in subsection \ref{subsec:NumericAlg}. We then perform an averaging process along with an alignment pooling strategy \cite{jia2012visual} to find a representative vector. Finally, we calculate the summation of this representative vector as the likelihood value. The candidate with the highest likelihood value is selected as the tracking result. We also update the templates throughout the sequence
using the same strategy as proposed in \cite{jia2012visual} to handle the appearance variations of the target region.

\subsection{Numerical Algorithm}\label{subsec:NumericAlg}

This section presents a numerical algorithm based on the ADMM \cite{boyd2011distributed} to efficiently solve the proposed model \eqref{eq:OurModel1}. The idea of the ADMM is to utilize auxiliary variables to convert a complicated convex problem to smaller sub-problems, where each one is efficiently solvable via an explicit formula. The ADMM iteratively solves the sub-problems until convergence. To do so, we first define vector $\mbf\in\Rbb^{k}$ such that $\mbf_{i} = \argmax{|\,\Cbf_{i}(:)|}$ and rewrite \eqref{eq:OurModel1} as: 
\begin{mini!} 
{\substack{\Cbf \in{\Rbb}^{{(lk)} \times {l}}\\ \mbf \in {\Rbb}^{k} }}{{\left\|\Xbf_j-\Dbf\Cbf\right\|}_{\mathrm{F}}^2+{\lambda}{{{\onebf}^{\top}_{k}}}{\mbf}} 
		{\label{eq:OurModel2}}{\label{eq:OurModel2_obj}}
        \addConstraint {\Cbf\geq 0} \label{eq:OurModel2_cons0}
		\addConstraint {\onebf^{\top}_{(lk)}} \Cbf = {\onebf^{\top}_{l}} \label{eq:OurModel2_cons1}
         \addConstraint {\mbf} \otimes {{\onebf}_{l}}{{\onebf}^{\top}_{l}} \geq \Cbf. \label{eq:OurModel2_cons2}
\end{mini!}
It should be noted that constraint \eqref{eq:OurModel2_cons2} is imposed in the above reformulation to ensure the equivalence between \eqref{eq:OurModel1} and \eqref{eq:OurModel2}. This inequality constraint can be transformed into an equality one by introducing a non-negative slack matrix $\Ubf\in\Rbb^{{(lk)} \times {l}}$, which compensates the difference between $ {\mbf} \otimes {{\onebf}_{l}}{{\onebf}^{\top}_{l}}$ and $\Cbf$. Using the resultant equality constraint, ${\onebf}^{\top}_{k}{\mbf}$ can be equivalently written as $\frac{1}{{l}^{2}}{{\onebf}^{\top}_{(lk)}}(\Cbf+\Ubf){{\onebf}_{l}}$. Moreover, this equality constraint implies that the columns of $\Cbf+\Ubf$ are regulated to be identical. Hence, one can simply replace it by a linear constraint independent of $\mbf$ as presented in \eqref{eq:OurModel4_cons2}. Therefore, we rewrite  \eqref{eq:OurModel2} independent of $\mbf$ as:

\begin{mini!} 
{\substack{\Cbf,\Ubf \in{\Rbb}^{{(lk)} \times {l}}}}{{\left\|\Xbf_j-\Dbf\Cbf\right\|}_{\mathrm{F}}^2+\!\frac{\lambda}{{l}^{2}}{{\onebf}^{\top}_{(lk)}}(\Cbf+\Ubf){{\onebf}_{l}}} 
		{\label{eq:OurModel4}}{\label{eq:OurModel4_obj}}
		\addConstraint {\Cbf\geq 0} \label{eq:OurModel4_cons0}
        \addConstraint {\onebf^{\top}_{(lk)}} \Cbf = {\onebf^{\top}_{l}} \label{eq:OurModel4_cons1}
         \addConstraint {\Ebf}(\Cbf\!+\!\Ubf)=\frac{\Ibf_{k}\otimes\onebf_{l}\!\onebft_{l}}{l}(\Cbf\!+\!\Ubf)\label{eq:OurModel4_cons2}
         \addConstraint  \Ubf\geq 0, \label{eq:OurModel4_cons3}
\end{mini!}
\begin{figure*}[t]
\centering
\includegraphics[width=\textwidth]{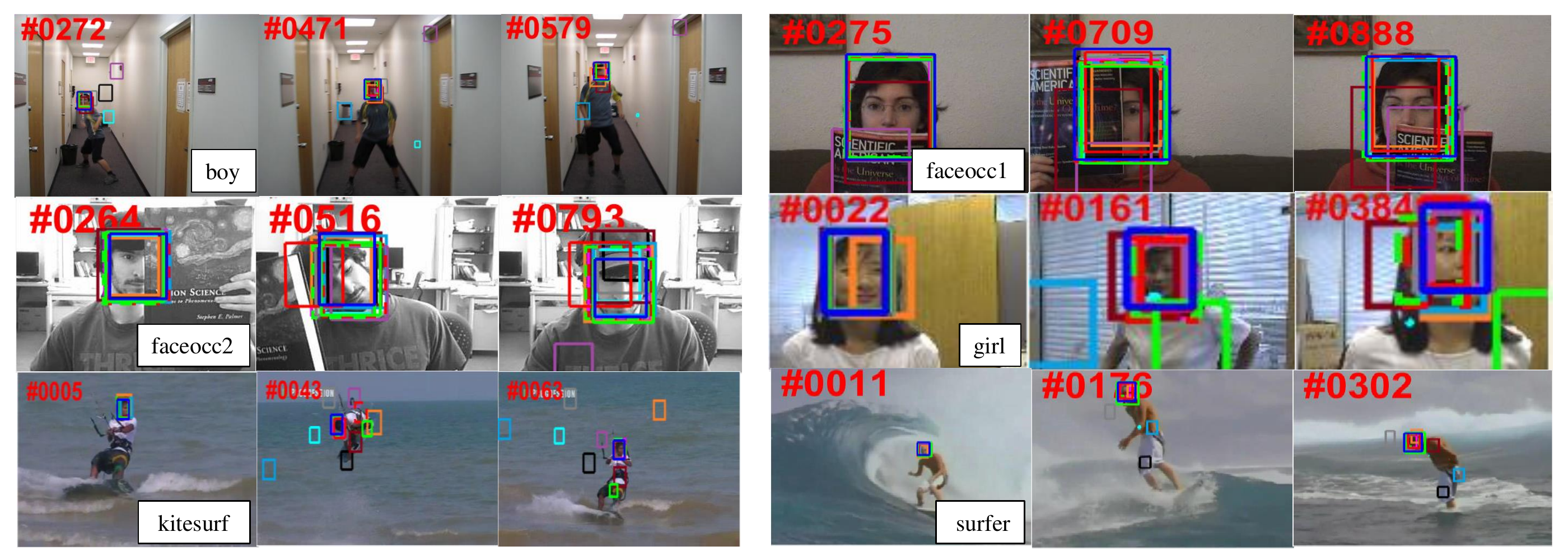}
\caption{Comparison of the tracking results of 11 state-of-the-art trackers and the two variants of the proposed SGLST on \textit{boy}, \textit{faceocc1}, \textit{faceocc2}, \textit{girl}, \textit{kitesurf}, and \textit{surfer} image sequences. Frame indices are shown at the top left corner of representative frames. Results are best viewed on high-resolution displays.\\
\small
 \centerline {(\textcolor{L1TColor}{\textbf{---}}L1T, \textcolor{StruckColor}{\textbf{---}}Struck, \textcolor{IVTColor}{\textbf{---}}IVT, \textcolor{MTTColor}{\textbf{---}}MTT, \textcolor{MILColor}{\textbf{---}}MIL, \textcolor{VTDColor}{\textbf{---}}VTD, \textcolor{FragColor}{\textbf{---}}Frag, \textcolor{ASLAColor}{\textbf{---}}ASLA, \textcolor{KCFColor}{\textbf{- - -}}KCF, \textcolor{MEEMColor}{\textbf{- - -}}MEEM,  \textcolor{RSST_HOGColor}{\textbf{---}}RSST\_HOG,  \textcolor{SGLST_ColorColor}{\textbf{---}}SGLST\_Color,  \textcolor{SGLST_HOGColor}{\textbf{---}}SGLST\_HOG)}}
 \label{fig:fig2-1}
\end{figure*}
\begin{figure*}[t]
\centering
\includegraphics[width=\textwidth]{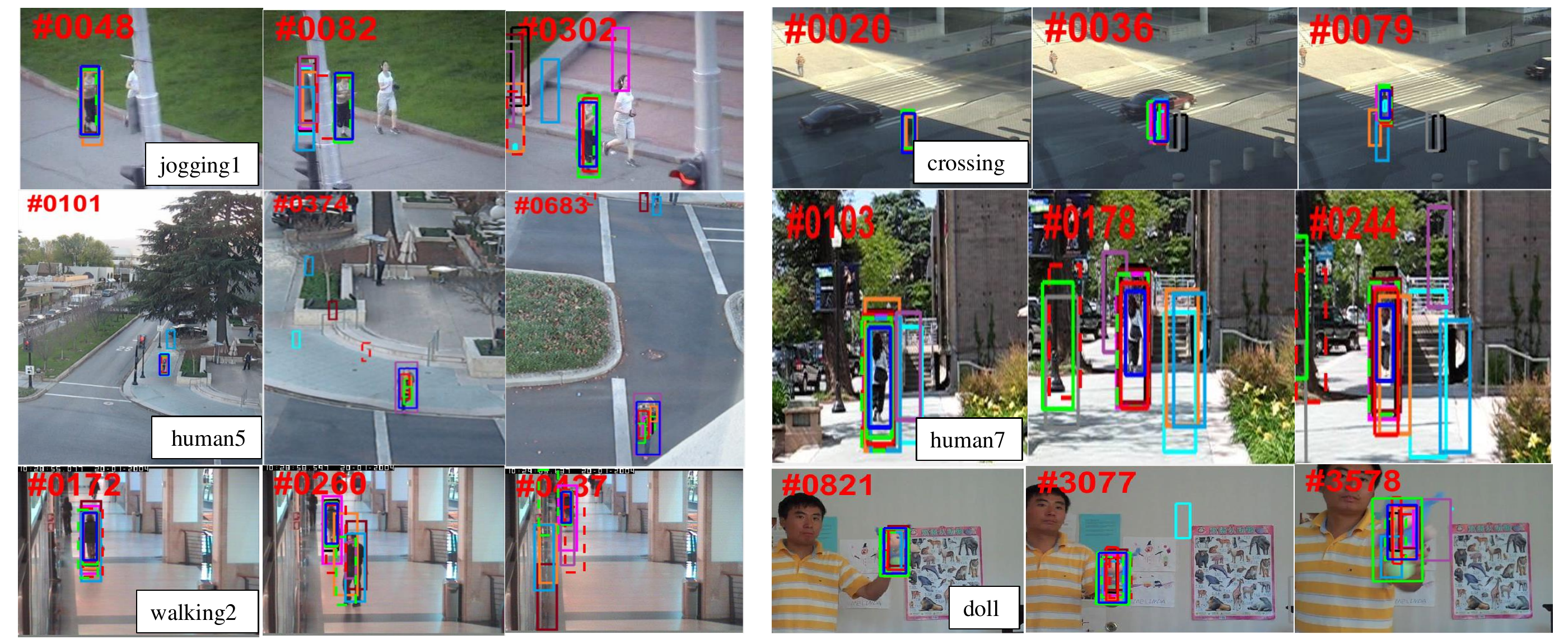}
\caption{Comparison of the tracking results of 11 state-of-the-art trackers and the two variants of the proposed SGLST on \textit{jogging1}, \textit{crossing}, \textit{human5}, \textit{human7}, \textit{walking2}, and \textit{doll} image sequences. Frame indices are shown at the top left corner of representative frames. Results are best viewed on high-resolution displays.\\
\small
 \centerline {(\textcolor{L1TColor}{\textbf{---}}L1T, \textcolor{StruckColor}{\textbf{---}}Struck, \textcolor{IVTColor}{\textbf{---}}IVT, \textcolor{MTTColor}{\textbf{---}}MTT, \textcolor{MILColor}{\textbf{---}}MIL, \textcolor{VTDColor}{\textbf{---}}VTD, \textcolor{FragColor}{\textbf{---}}Frag, \textcolor{ASLAColor}{\textbf{---}}ASLA, \textcolor{KCFColor}{\textbf{- - -}}KCF, \textcolor{MEEMColor}{\textbf{- - -}}MEEM,  \textcolor{RSST_HOGColor}{\textbf{---}}RSST\_HOG,  \textcolor{SGLST_ColorColor}{\textbf{---}}SGLST\_Color,  \textcolor{SGLST_HOGColor}{\textbf{---}}SGLST\_HOG)}}
 \label{fig:fig2-2}
\end{figure*}
where matrix $\Ebf$ serves as the right circular shift operator on the rows of $\Cbf+\Ubf$. To construct the ADMM formulation, whose subproblems possess closed-form solutions, we define auxiliary variables $\Cbfhat,\Ubfhat\in\Rbb^{{(lk)} \times {l}}$ and reformulate \eqref{eq:OurModel4} as:
\begin{subequations} \label{eq:OurModel5}
\begin{align}
&\!\!\!\!\!\underset{\begin{subarray}{l} {\Cbf,\Cbfhat,\Ubf,\Ubfhat \in{\Rbb}^{{(lk)} \times {l}}} \end{subarray}}
{\!\!\!\!\text{minimize}} &&\!\!\!\!\!\!{\left\|\Xbf_j-\Dbf\Cbf\right\|}_{\mathrm{F}}^2+\!\frac{\lambda}{{l}^{2}}{{\onebf}^{\top}_{(lk)}}(\Cbf+\Ubf){{\onebf}_{l}}  \nonumber\\ 
&&&\!\!\!\!\!\!+\frac{{\mu}_1}{2}{\left\|\Cbf-{\Cbfhat}\right\|}_{\mathrm{F}}^2+\frac{{\mu}_2}{2}{\left\|\Ubf-{\Ubfhat}\right\|}_{\mathrm{F}}^2 \label{eq:OurModel5_obj}\\
&{\text{subject to}}
&& \!\!\!\!\!\! \Cbfhat \geq 0, \label{eq:OurModel5_cons0}\\
&&&\!\!\!\!\!\! {\onebf^{\top}_{(lk)}} \Cbfhat = {\onebf^{\top}_{l}}\, \label{eq:OurModel5_cons1}\\
&&&\!\!\!\!\!\! {\Ebf}(\Cbf\!+\!\Ubf)=\frac{\Ibf_{k}\otimes\onebf_{l}\!\onebft_{l}}{l}(\Cbf\!+\!\Ubf),\label{eq:OurModel5_cons3}\\
&&&\!\!\!\!\!\! \Ubfhat \geq 0, \label{eq:OurModel5_cons4}\\
&&&\!\!\!\!\!\! \Cbf=\Cbfhat,~~ \Ubf=\Ubfhat. \label{eq:OurModel5_cons5} 
\end{align}
\end{subequations}
where $\mu_1,\mu_2>0$ are the augmented Lagrangian parameters. Without loss of generality, we assume $\mu=\mu_1=\mu_2$ \cite{boyd2011distributed}. The last two terms in the objective function \eqref{eq:OurModel5_obj} are then vanished for any feasible solutions, which implies \eqref{eq:OurModel4} and \eqref{eq:OurModel5} are equivalent. We further form the augmented Lagrangian function to solve \eqref{eq:OurModel5} as follows:
\begin{equation}
\begin{aligned}
&\mathcal{L}_{\mu}\!(\Cbf,\msh\Ubf,\msh\Cbfhat,\msh\Ubfhat,\msh{\Lambdabf}_1,\msh{\Lambdabf}_2)\msh=\!
&&\!\!\!{\left\|\Xbf_j\msh-\msh\Dbf\Cbf\right\|}_{\mathrm{F}}^2\msh+\msh\frac{\lambda}{{l}^{2}}{{\onebf}^{\top}_{(lk)}}\msh(\Cbf\msh+\msh\Ubf){\onebf_l}\\
&&&  \hspace{-2.5cm} +\msh\frac{\mu}{2}{\left\|\Cbf\msh-\msh{\Cbfhat}\msh+\msh\frac{{\Lambdabf}_1}{\mu}\right\|}_{\mathrm{F}}^2
\!\msh+\!\frac{{\mu}}{2}{\left\|\Ubf\msh-\msh\Ubfhat\msh+\msh\frac{{\Lambdabf}_2}{\mu}\right\|}_{\mathrm{F}}^2\label{eq:OurModel6} 
\end{aligned}\vspace{0.5cm}
\end{equation}
where $\bm \Lambda_{1},\Lambdabf_{2}\in\Rbb^{{(lk)} \times {l}}$ are the Lagrangian multipliers corresponding to the equations in \eqref{eq:OurModel5_cons5}. 

Given initialization for $\Cbfhat$, $\Ubfhat$, $\Lambdabf_{1}$, and $\Lambdabf_{2}$ at time $t=0$ (e.g., ${\Cbfhat}^0,{\Ubfhat}^0,{\Lambdabf}_1^0,{\Lambdabf}_2^0$),  \eqref{eq:OurModel6} is solved through the ADMM iterations.  At the next iteration, $\Cbf$ and $\Ubf$ are updated by minimizing \eqref{eq:OurModel6} under the constraint \eqref{eq:OurModel5_cons3}. To do so, we first define $\{\zbf_i\}_{i=1}^{lk}$, where $\zbf_i\in\Rbb^{2l}$ is obtained by stacking the $i^{th}$ rows of $\Cbf$ and $\Ubf$. We then divide this minimization problem into $lk$ equality constrained quadratic programs, where each program has its analytical solution. Using the updated $\Cbf$ and $\Ubf$, we compute ${\Cbfhat}$ and ${\Ubfhat}$ by minimizing \eqref{eq:OurModel6} with the constraints \eqref{eq:OurModel5_cons0},~\eqref{eq:OurModel5_cons1},~\eqref{eq:OurModel5_cons4}. To this end, we split the problem into two separate subproblems with closed-form solutions over $\Cbfhat$ and $\Ubfhat$, where the first subproblem consists of $l$ independent Euclidean norm projections onto the probability simplex constraints and the second subproblem consists of $l$ independent Euclidean norm projections onto the non-negative orthant. Finally, we update $\Lambdabf_1$ and $\Lambdabf_2$ by performing $l$ parallel updates over their respective columns. All these iterative updates can be quickly performed  due to the closed-form solutions.

\section{Experimental Results}\label{sec:exp}
In this section, we evaluate the performance of the proposed SGLST on 16 publicly available frame sequences, the OTB50 \cite{wu2013online}, and the OTB100 \cite{wu2015object} tracking benchmarks.

We resize each target region to $32\times 32$ pixels and extract overlapping local patches of $16 \times 16$ pixels inside the target region using the step size of 8 pixels. This leads to $l=9$ local patches. For each local patch, we extract two sets of features, namely, gray-level intensity features and histogram of oriented gradients (HOG) features, to represent its characteristics from two perspectives. Both features have shown promising tracking results in different trackers and HOG features \cite{dalal2005histograms} have demonstrated significant improvement in visual tracking \cite{henriques2012exploiting,mei2015robust,zhang2018robust}. The proposed SGLST therefore has two variants: SGLST\_Color and SGLST\_HOG. 
For the HOG features, we resize the target candidates to $64\times 64$ pixels and exploit 196 dimensional HOG features for each of the $32 \times 32$ local patches to capture relatively high-resolution edge information. For all the experiments, we set $\lambda = 0.1$, $\mu_1=\mu_2=\mu=0.1$, the number of particles $n=400$, and the number of target templates $k=10$. We adopt the same setting as used in \cite{jia2012visual} to update templates.

\begin{figure*}[t]
\centering
\includegraphics[width=\textwidth]{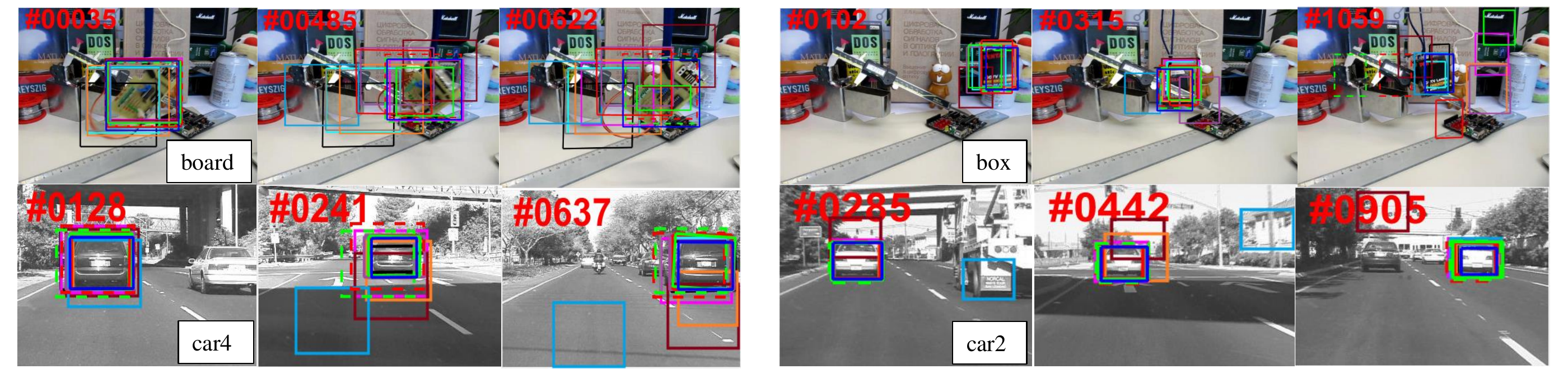}
\caption{Comparison of the tracking results of 11 state-of-the-art trackers and the two variants of the proposed SGLST on \textit{board}, \textit{box}, \textit{car4}, and \textit{car2} image sequences. Frame indices are shown at the top left corner of representative frames. Results are best viewed on high-resolution displays.\\
\small
 \centerline {(\textcolor{L1TColor}{\textbf{---}}L1T, \textcolor{StruckColor}{\textbf{---}}Struck, \textcolor{IVTColor}{\textbf{---}}IVT, \textcolor{MTTColor}{\textbf{---}}MTT, \textcolor{MILColor}{\textbf{---}}MIL, \textcolor{VTDColor}{\textbf{---}}VTD, \textcolor{FragColor}{\textbf{---}}Frag, \textcolor{ASLAColor}{\textbf{---}}ASLA, \textcolor{KCFColor}{\textbf{- - -}}KCF, \textcolor{MEEMColor}{\textbf{- - -}}MEEM,  \textcolor{RSST_HOGColor}{\textbf{---}}RSST\_HOG,  \textcolor{SGLST_ColorColor}{\textbf{---}}SGLST\_Color,  \textcolor{SGLST_HOGColor}{\textbf{---}}SGLST\_HOG)}}
 \label{fig:fig2-3}
\end{figure*}
\subsection{Experimental Results on Publicly Available Sequences}
We conduct extensive experiments on 16 challenging frame sequences and compare SGLST\_Color and SGLST\_HOG with 11 state-of-the-art trackers, namely, L1T \cite{mei2011robust}, Struck \cite{hare2016struck}, IVT \cite{ross2008incremental}, MTT  \cite{zhang2012robust}, MIL \cite{babenko2011robust},  VTD \cite{kwon2010visual}, Frag \cite{adam2006robust}, ASLA \cite{jia2012visual}, KCF \cite{henriques2015high}, MEEM \cite{zhang2014meem}, and RSST\_HOG \cite{zhang2018robust}. To ensure fair comparison, we use the available source code or the binary code together with the optimal parameters provided by the respective authors to produce the tracking results.

Figure \ref{fig:fig2-1}, Figure \ref{fig:fig2-2}, and Figure \ref{fig:fig2-3} demonstrate the tracking results of the 13 aforementioned compared methods on three representative frames of each of the 16 sequences. 
\begin{table*}[t]
\footnotesize
\begin{center}
\begin{tabular}{l|c|c|c|c|c|c|c|c|c|c|c|c|c}
\hline
Seq & L1T&Struck&IVT&MTT&MIL&VTD&Frag&ASLA&KCF&MEEM&RSST\_HOG&SGLST\_Color&SGLST\_HOG \\
\hline\hline
boy   &0.73  &0.76&0.26&0.49&0.49&0.62&0.38&0.36&0.77& \textcolor{red}{0.79} & 0.76& \textcolor{blue}{\textbf{0.81}}  & 0.78  \\
faceocc1&0.74&0.73&0.72&0.70&0.60&0.69&0.82&0.41& \textcolor{red}{0.76}&0.75&0.70&0.74& \textcolor{blue}{\textbf{0.78}}\\
faceocc2&0.68& \textcolor{red}{0.76}&0.72&0.74&0.67 &  0.71 & 0.65& 0.65 & 0.74&  \textcolor{blue}{\textbf{0.78}}& 0.67& 0.75 &0.72 \\
girl & 0.73& \textcolor{red}{0.74} &  0.16  & 0.66& 0.39 &0.60& 0.43 & 0.72&0.58 &   0.69 &  \textcolor{blue}{\textbf{ 0.75}} &  0.21 & 0.70   \\
kitesurf &   0.25& 0.64& 0.30& 0.35& 0.31& 0.15& 0.19& 0.25& 0.48& \textcolor{red}{0.67}& \textcolor{blue}{\textbf{0.71}}& 0.22& 0.47\\
surfer &   0.04&0.41 & 0.06&0.10&0.25&0.31&0.22&0.41&0.47&0.51&0.61& \textcolor{blue}{\textbf{0.71}}& \textcolor{red}{0.65}\\
jogging1 &  0.14 &0.17 &0.17 &0.17   &0.18 &0.21   &0.52 &  0.22  & 0.25  & 0.67   & 0.71   & \textcolor{blue}{\textbf{0.78}}  & \textcolor{red}{0.74} \\
crossing&0.24&0.67&0.29&0.19&0.72&0.31&0.31& \textcolor{red}{0.77}&0.71&0.71&0.76&0.72& \textcolor{blue}{\textbf{0.79}}\\
human5 &   0.38& 0.35& 0.18& 0.45& 0.21& 0.28& 0.03& \textcolor{red}{0.68}& 0.21& 0.28& 0.51& 0.35& \textcolor{blue}{\textbf{0.72}}\\
human7 &    0.51&0.48&0.23&0.28&0.48&0.28&0.27&0.29&0.29&0.48&\textcolor{red}{0.58}&0.40& \textcolor{blue}{\textbf{0.83}} \\
walking2 &   0.75 &0.51 &0.79 &0.78&0.29&0.40&0.35&0.37&0.39&0.31&0.76& \textcolor{red}{0.80}& \textcolor{blue}{\textbf{0.83}} \\
doll&0.46 & 0.55 &0.43&0.39 &0.42&0.66&0.61& \textcolor{red}{0.78}&0.59&0.60&0.48& \textcolor{blue}{\textbf{0.84}}&0.75 \\
board &  0.13&0.66&0.19&0.19&0.40&0.28&0.52&0.30 &0.65 & \textcolor{red}{0.68} &0.42 &0.56& \textcolor{blue}{\textbf{0.75}} \\
box &  \textcolor{red}{0.55}&0.21&0.51&0.21&0.27&0.42&0.46&0.34 &0.35 &0.31 &0.33&0.21& \textcolor{blue}{\textbf{0.62}}\\
car4 &0.72&0.48&0.82&0.75&0.25&0.36&0.19&0.75&0.48&0.45& \textcolor{blue}{\textbf{0.87}}&0.82 & \textcolor{red}{0.85}\\
car2 &0.86 &0.68& 0.89 & 0.87 & 0.16 &0.80&0.25&0.85&0.68&0.68& \textcolor{blue}{\textbf{0.91}} &0.81& \textcolor{red}{0.88} \\
\hline
Average  &  0.49  &  0.55  &  0.42  & 0.46   &  0.38  &  0.44  &  0.39  &   0.51 &  0.53  &  0.59  &  \textcolor{red}{0.66}  &  0.61  &  \textcolor{blue}{\textbf{0.74}}  \\
\hline
\end{tabular}
\end{center}
\caption{Summary of the average overlap scores of 13 compared methods on 16 sequences. The bold numbers in \textcolor{blue}{blue} indicate the best performance, while the numbers in  \textcolor{red}{red} indicate the second best.}
\label{tab:tab1}
\end{table*}
Here, we briefly analyze the tracking performance of each compared tracker under different challenging scenarios. The L1T tracker fails when the target undergoes fast motion and rotation as shown in \textit{Kitesurf} and \textit{surfer} sequences, occlusion as shown in the \textit{jogging1} sequence, or scale variation as shown in the \textit{board} sequence. Struck cannot track the target when occlusion (\textit{jogging1} and \textit{box}) or fast motion (\textit{surfer}) occurs. IVT drifts from the target in the frame sequences containing the out-of-view challenge (\textit{girl} and \textit{jogging}), fast motion (\textit{boy}), or scale variation (\textit{human5}). MTT loses the target having large motions between consecutive frames (\textit{board} and \textit{crossing}). MIL fails to track the target when scale variation (\textit{car4} and \textit{car2}) or occlusion (\textit{walking2}) happens. VTD and Frag lead to the drift of the target under fast motion and deformation circumstances as shown in \textit{crossing} and \textit{human7} sequences. In addition, they cannot adequately handle scale variation as shown in the \textit{box} sequence. ASLA does not yield good performance in the cases of heavy occlusions (\textit{faceocc1}, \textit{jogging}, and \textit{walking2}). KCF is incapable of dealing with scale variation (\textit{car4} and \textit{walking2}), occlusion \textit {(jogging1)}, or out-of-view challenges (\textit{box}). MEEM achieves good overall performance.  However, it drifts from the target when scale varies (\textit{car4}) and does not sufficiently address the challenge of partial occlusion (\textit{walking2} and \textit{box}). RSST\_HOG performs well in  most sequences, but it drifts away in the sequences with scale variations (\textit{doll}, \textit{board}, and \textit{box}). SGLST\_Color also demonstrates favorable performance in most of the sequences. However, it encounters problems when illumination changes happen (\textit{kitesurf} and \textit{box}). Among all the compared methods, SGLST\_HOG performs well in tracking human faces, human bodies, objects, and vehicles in the 16 challenging sequences. The favorable performance of the proposed SGLST reflects the advantages of adopting local patches within the target and keeping the spatial structure among local patches. In addition, using HOG features in SGLST helps to improve the tracking performance yielded by using intensity features.

For quantitative comparison, we compute the average overlap score across all frames of each image sequence for each compared method. It is worthy of mentioning that the overlap score between the tracked bounding box $r_t$ and the ground truth bounding box $r_g$ is defined as $S=\frac{|r_t\cap r_g|}{|r_t\cup r_g|}$, where $|\cdot|$ is the number of pixels in the bounding box, $\cap$ represents the intersection of the two bounding boxes, and $\cup$ represents the union of the two bounding boxes. Table \ref{tab:tab1} summarizes the average overlap scores across all frames of each of 16 sequences for compared methods. It is clear that the two proposed trackers, SGLST\_Color and SGLST\_HOG, achieve overall favorable tracking performance for the tested sequences. On average, SGLST\_Color drastically improves the average overlap scores of L1T, IVT, MTT, MIL, VTD, and Frag by 24.49\%,  45.24\%, 32.61\%, 60.53\%, 38.64\%, and 56.41\%, respectively. It also outperforms Struck, ASLA, KCF, and MEEM by improving their average overlap scores by 10.91\%, 19.61\%, 15.09\%, and 3.39\%, respectively. RSST\_HOG is the only tracker that outperforms SGLST\_Color by 8.2\% mainly due to the use of HOG features. The proposed SGLST\_HOG achieves the best average overlap score and significantly outperforms SGLST\_Color and RSST\_HOG by 21.31\% and 12.12\%, respectively. In summary, the qualitative results shown in Figure \ref{fig:fig2-1}, Figure \ref{fig:fig2-2}, and Figure \ref{fig:fig2-3} and the quantitative results shown in Table \ref{tab:tab1} demonstrate that SGLST\_HOG achieves the best tracking performance and SGLST\_Color achieves the third best tracking performance, inferior to RSST\_HOG that uses HOG features instead of intensity features. Both variants of the proposed SGLST can successfully track the targets in a majority of frames in all 16 tested sequences with different challenging conditions such as fast motion, rotation and scale variations, occlusions, and illumination changes.  


\begin{figure*}[t]
\centering     
\subfigure{\includegraphics[width=0.33\textwidth]{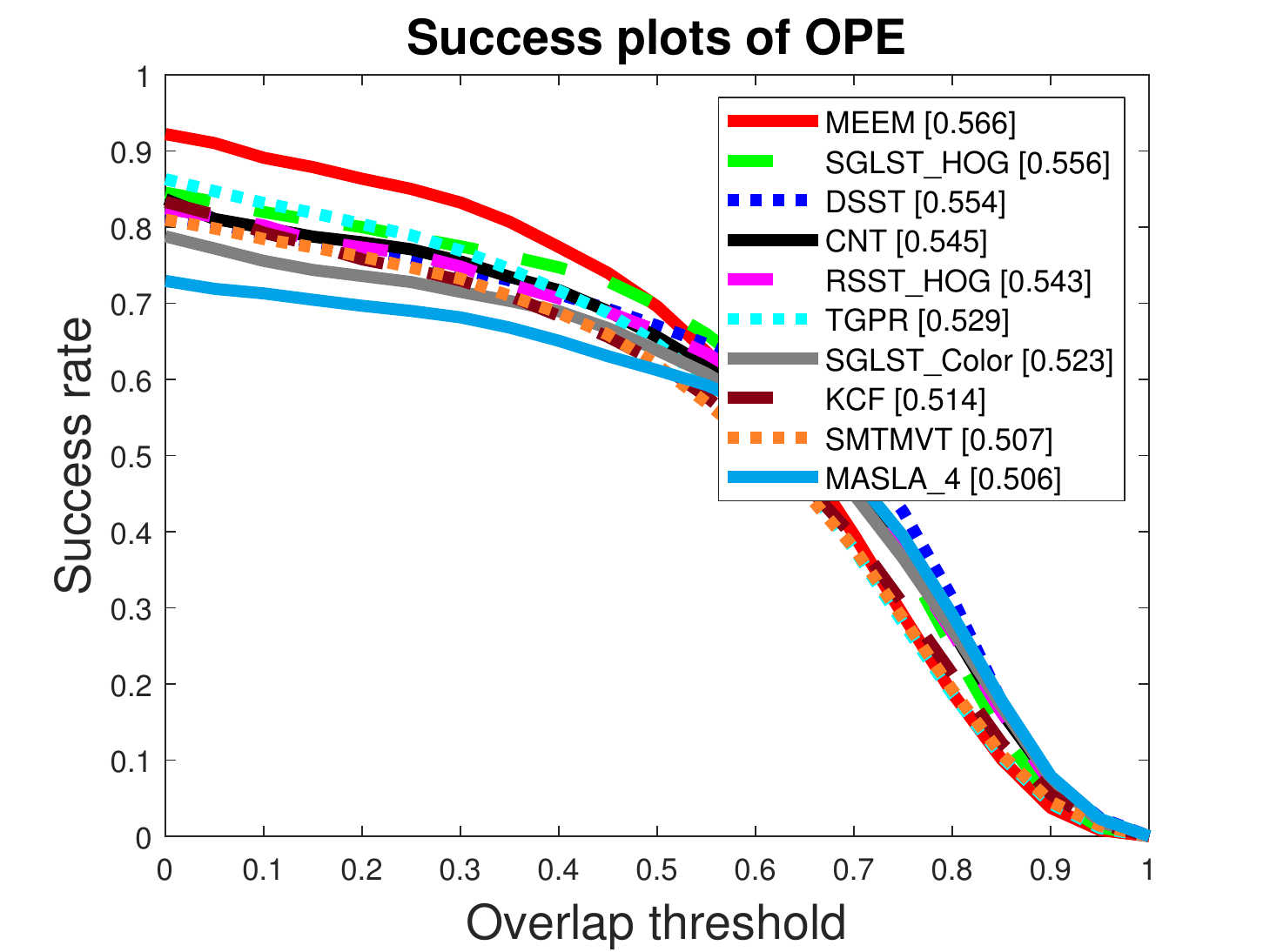}}
\subfigure{\includegraphics[width=0.33\textwidth]{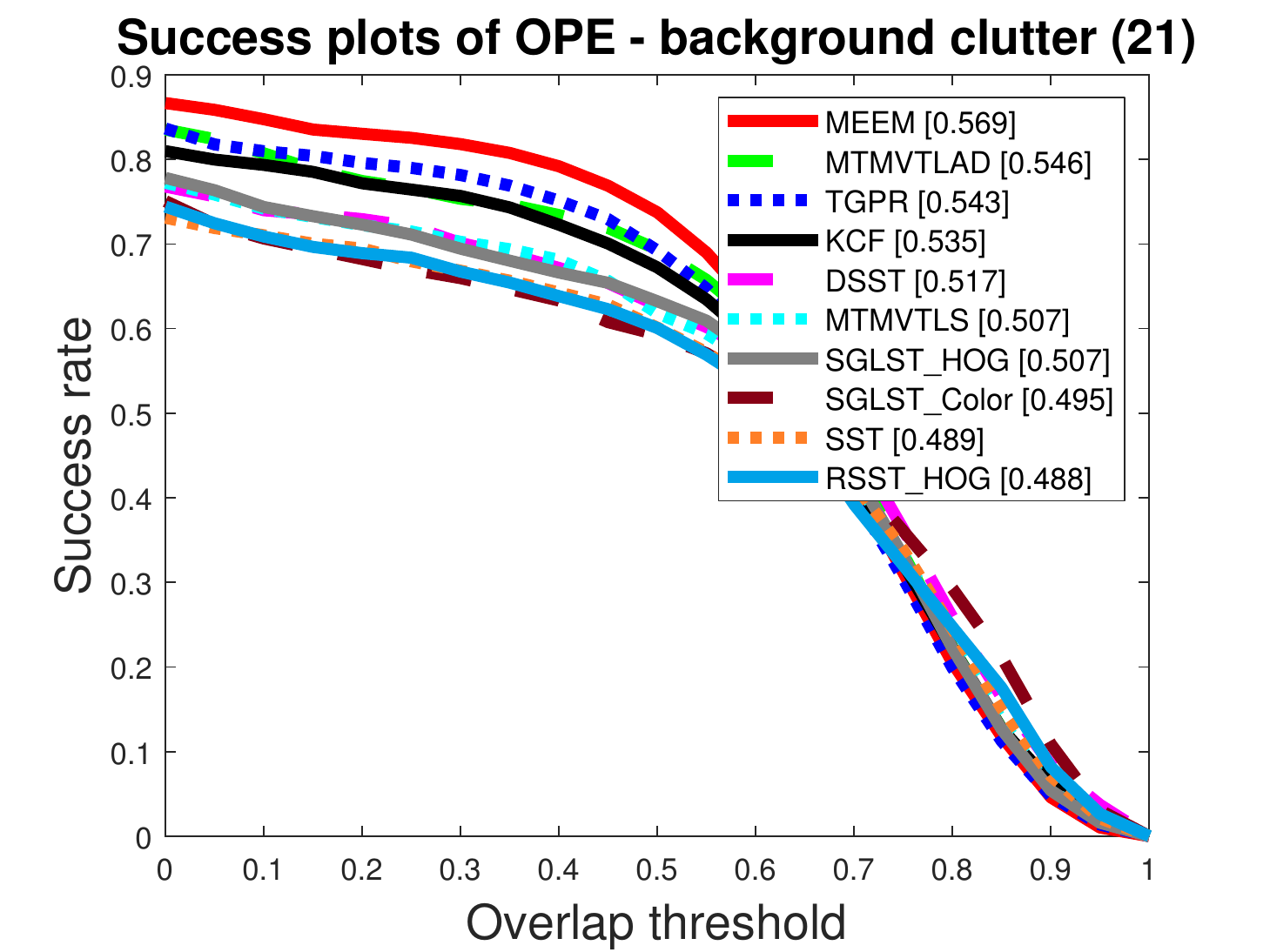}}
\subfigure{\includegraphics[width=0.33\textwidth]{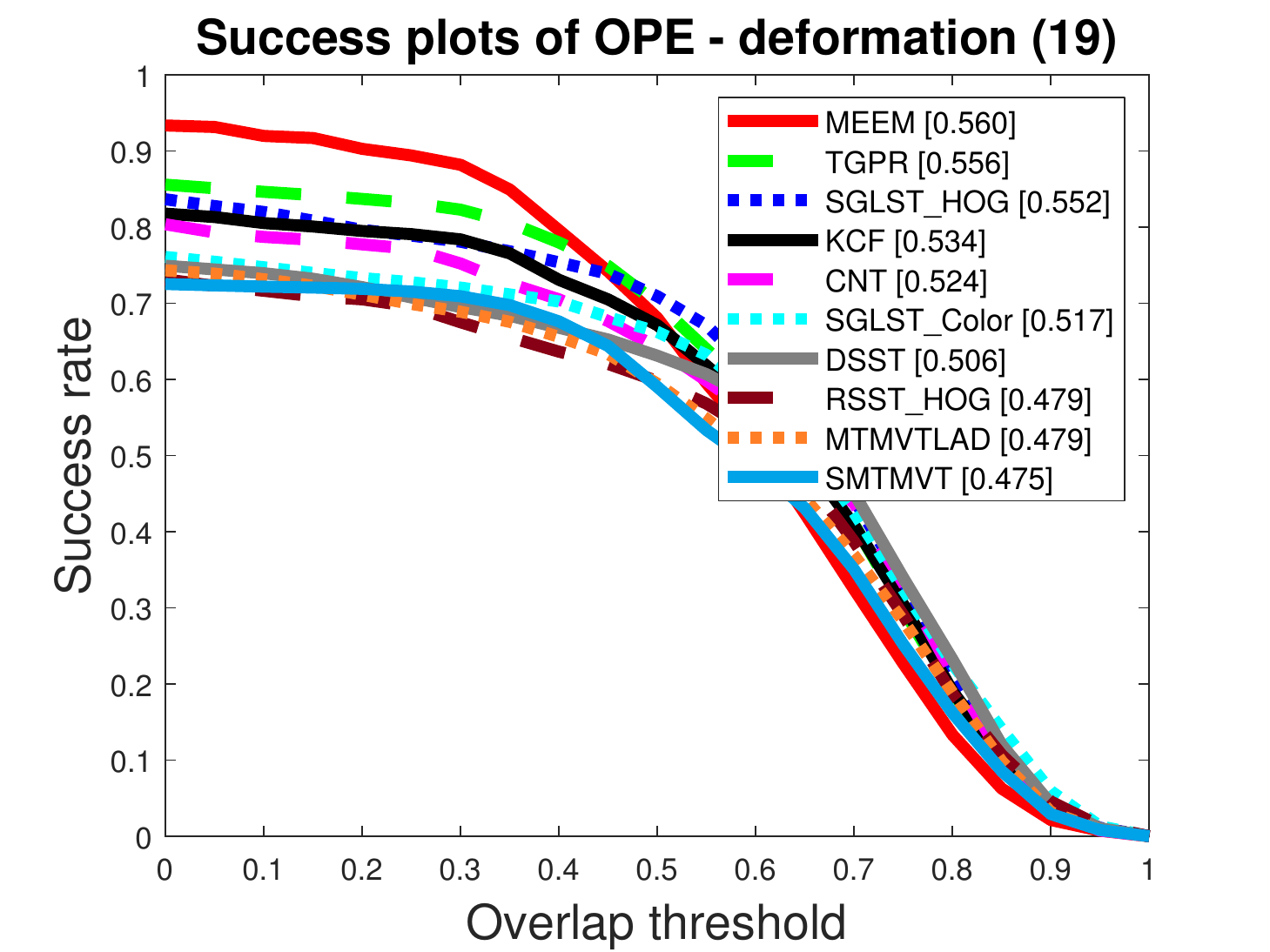}}
\subfigure{\includegraphics[width=0.33\textwidth]{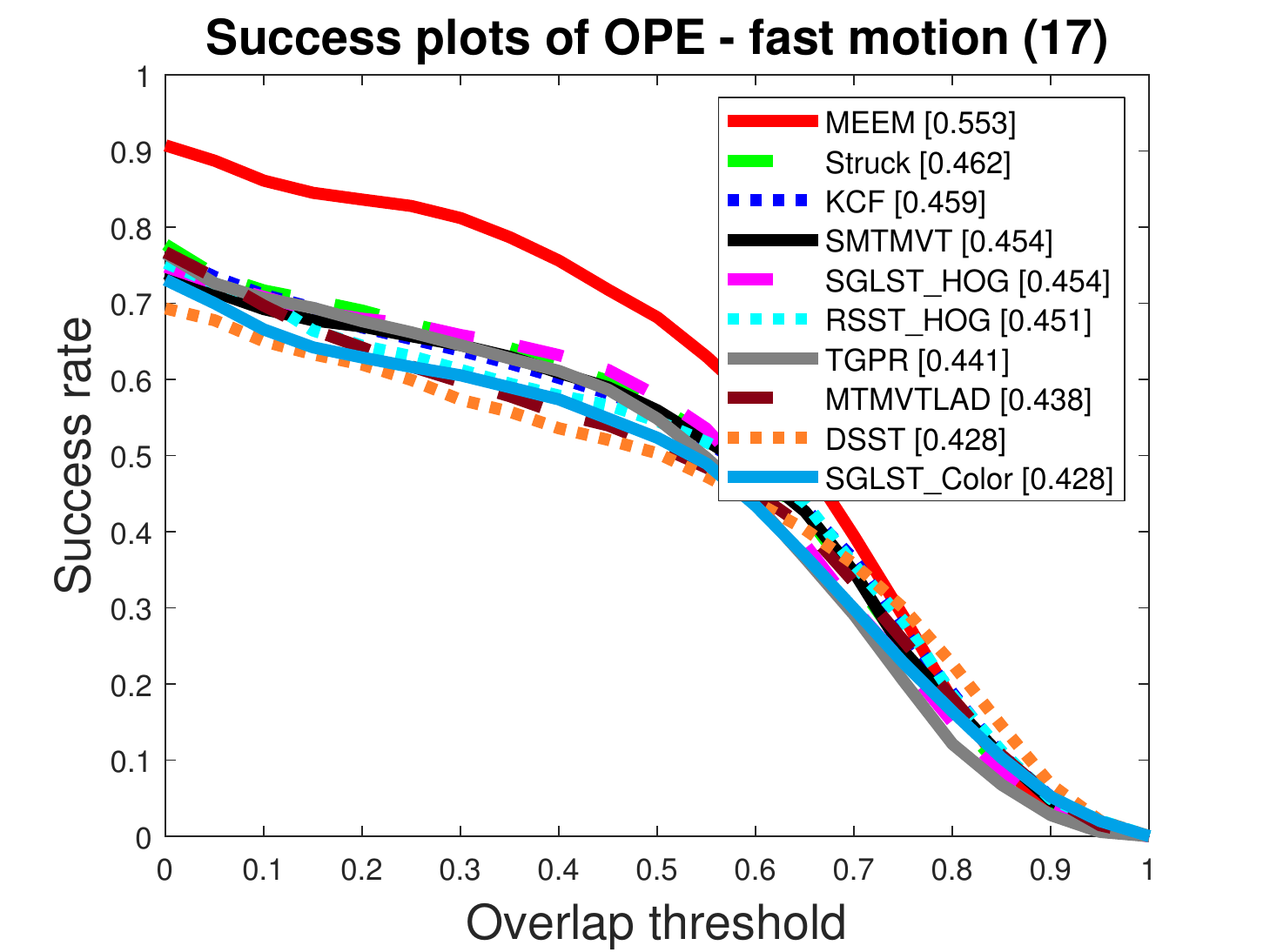}}
\subfigure{\includegraphics[width=0.33\textwidth]{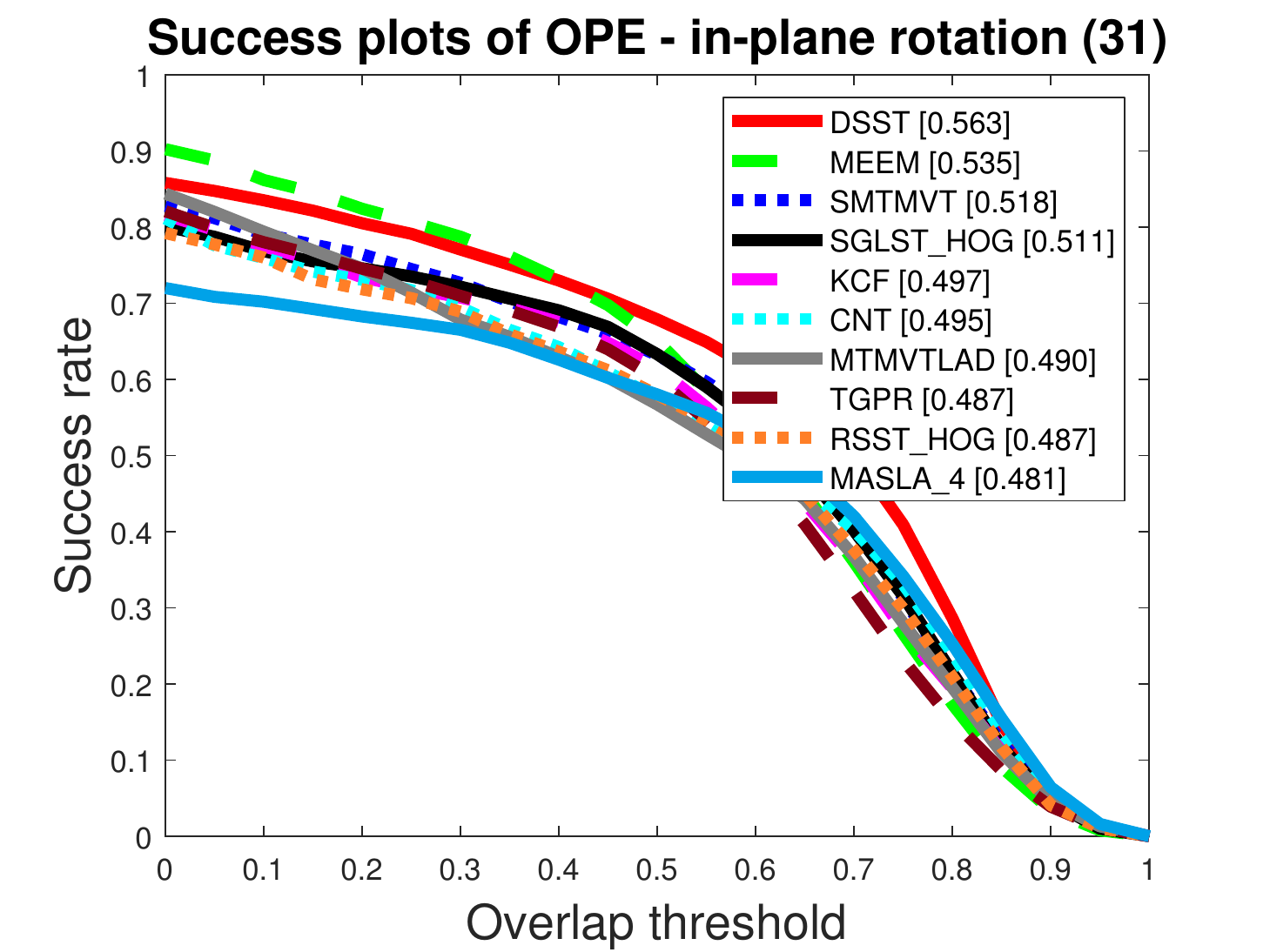}}
\subfigure{\includegraphics[width=0.33\textwidth]{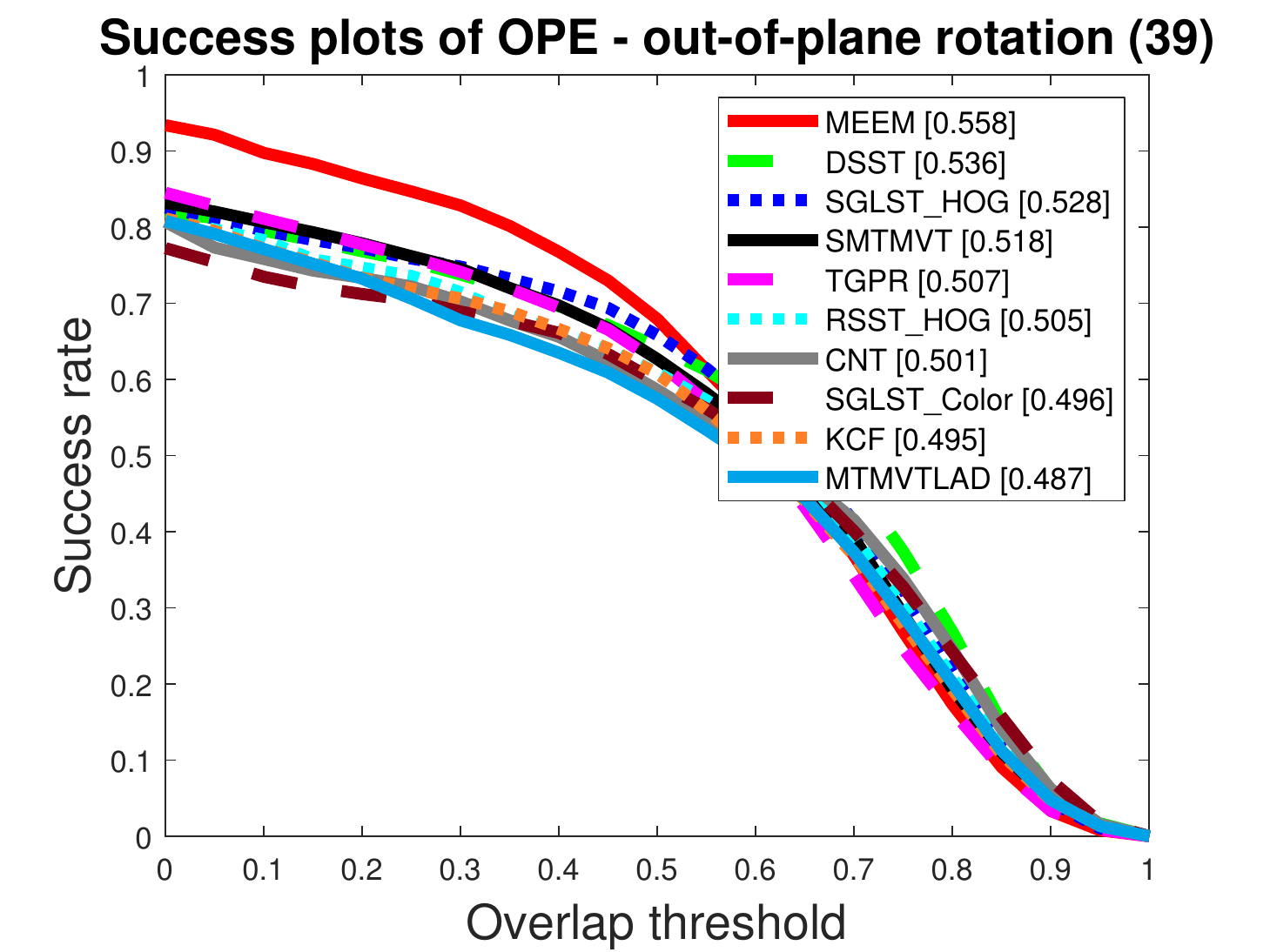}}
\caption{OTB50 overall OPE success plots and the OPE success plot BC, DEF, FM, IPR, and OPR challenge subsets. The value appearing in the title is the number of sequences in the specific subset. The values appearing in the legend are the AUC scores. Only the top 10 trackers are presented, while the results of the other trackers can be found in \cite{wu2013online}.}
\label{fig:fig3-1}
\end{figure*}
\begin{figure*}[t]
\centering     
\subfigure{\includegraphics[width=0.33\textwidth]{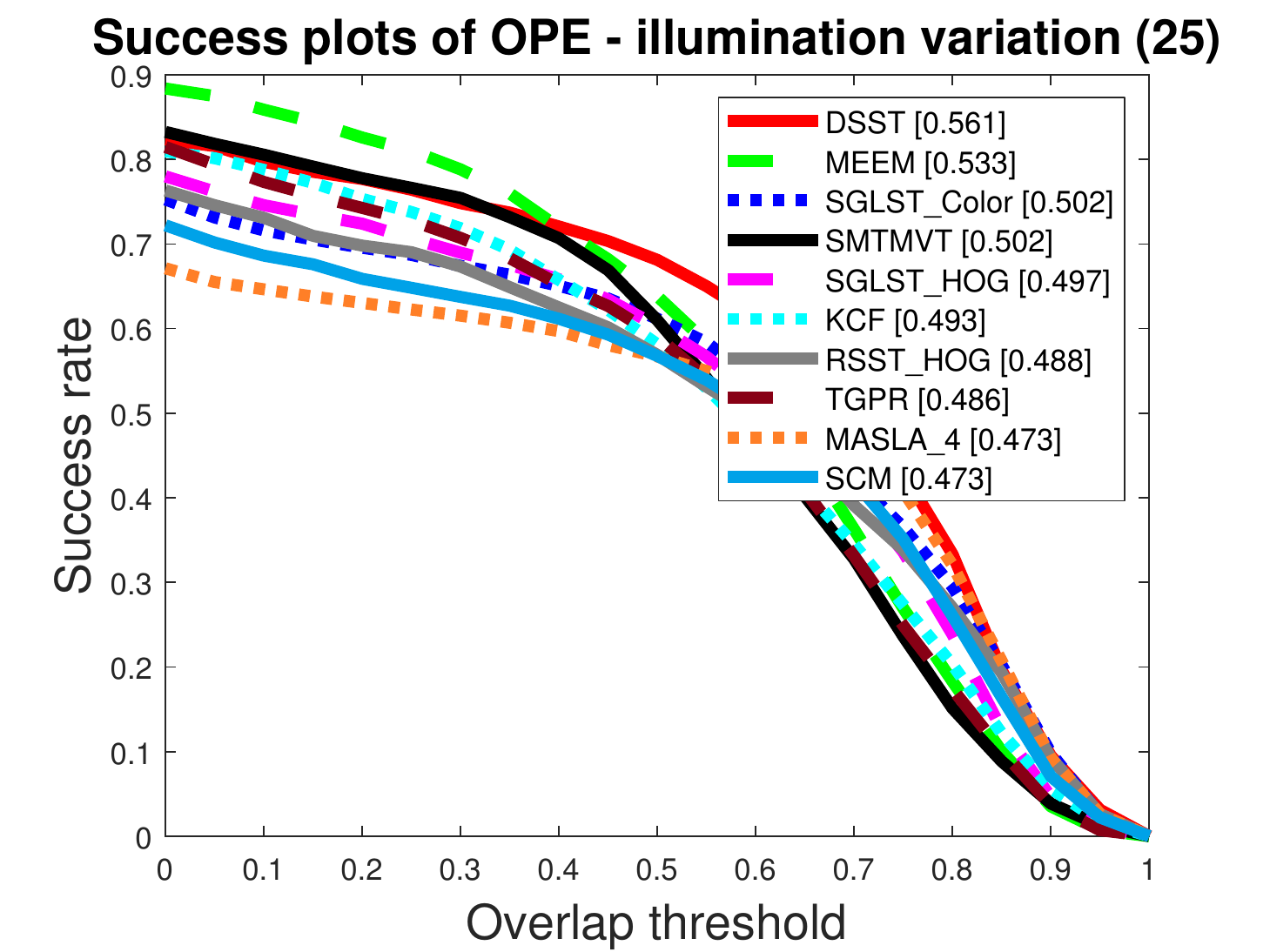}}
\subfigure{\includegraphics[width=0.33\textwidth]{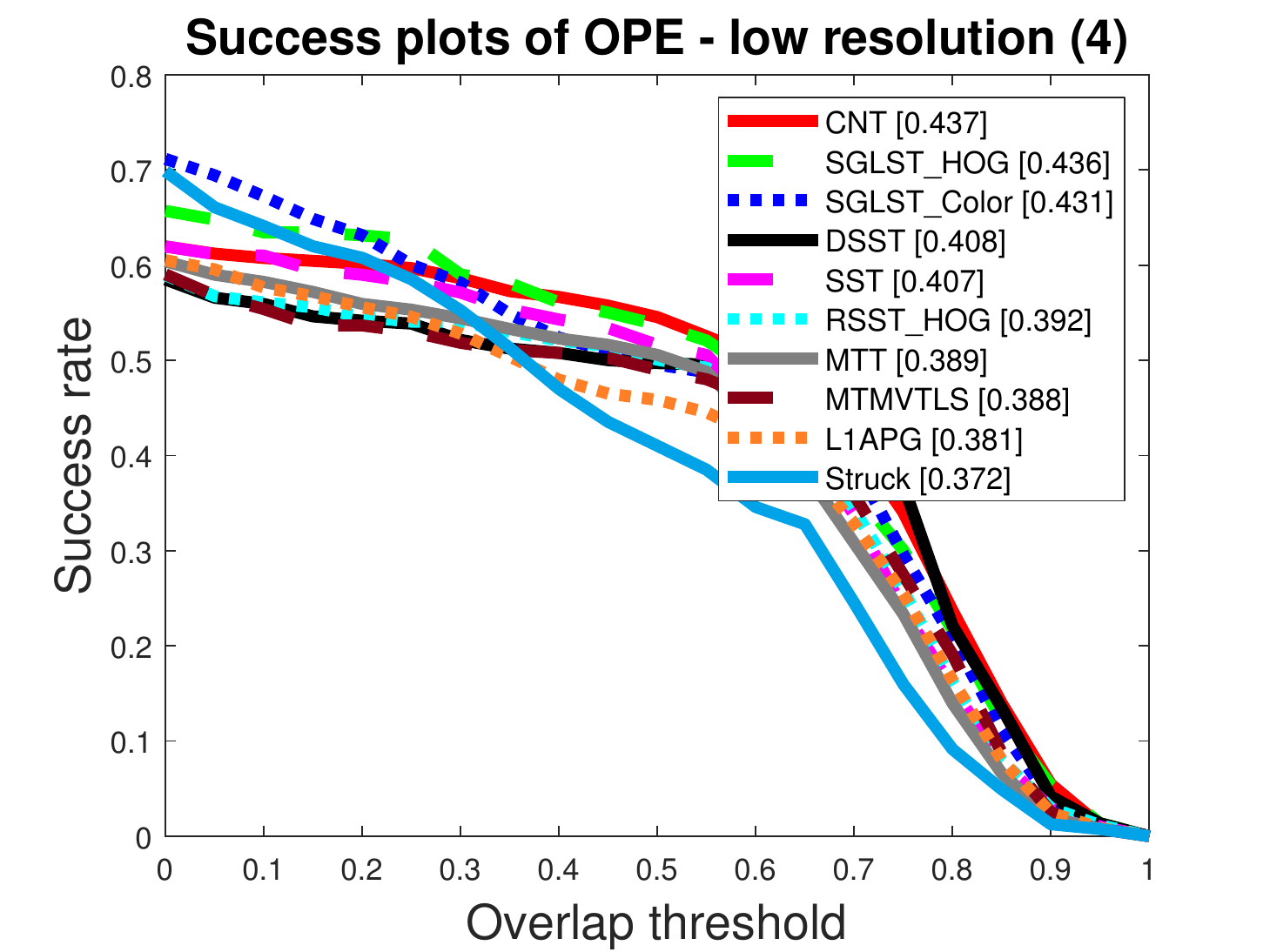}}
\subfigure{\includegraphics[width=0.33\textwidth]{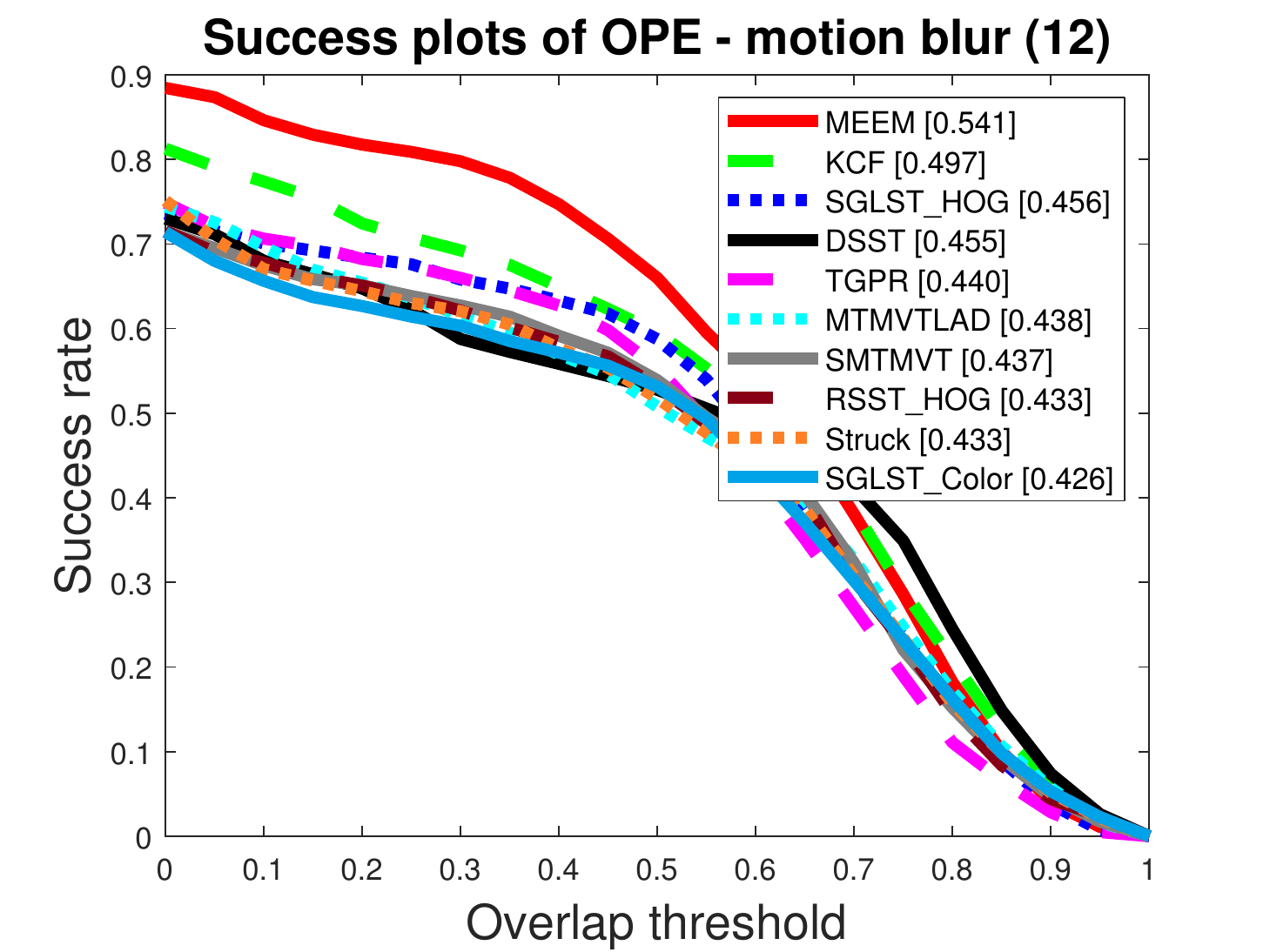}}
\subfigure{\includegraphics[width=0.33\textwidth]{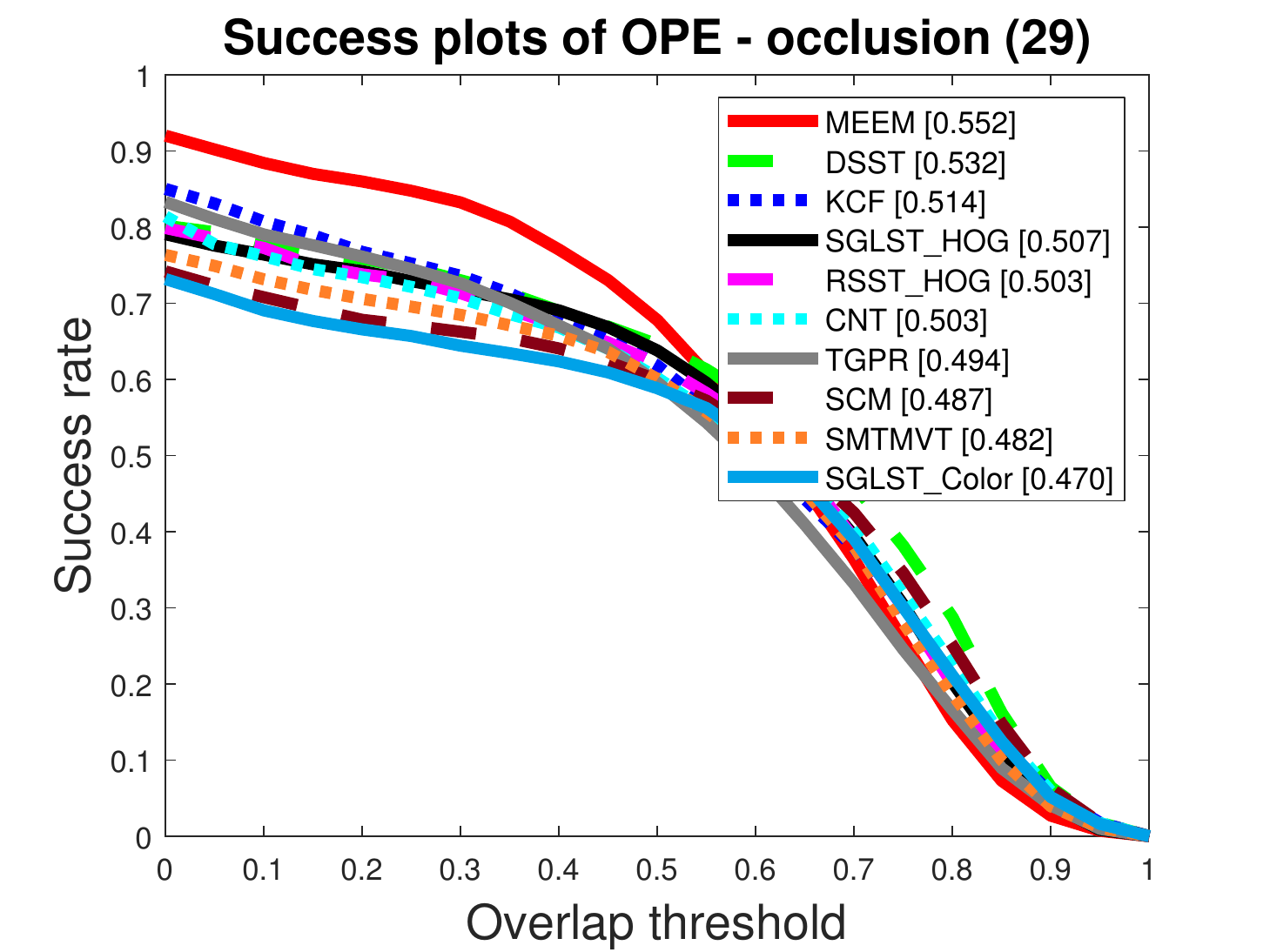}}
\subfigure{\includegraphics[width=0.33\textwidth]{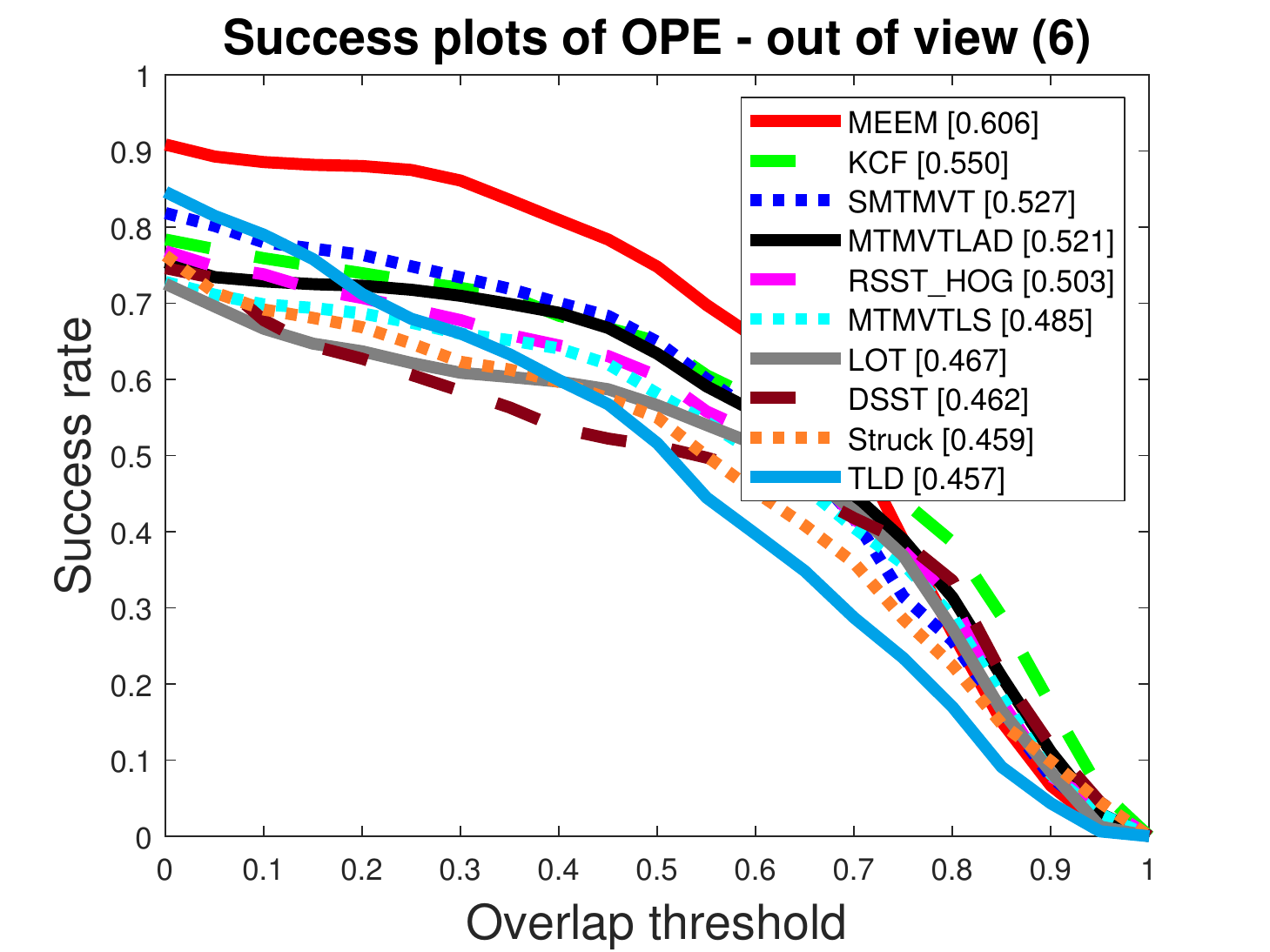}}
\subfigure{\includegraphics[width=0.33\textwidth]{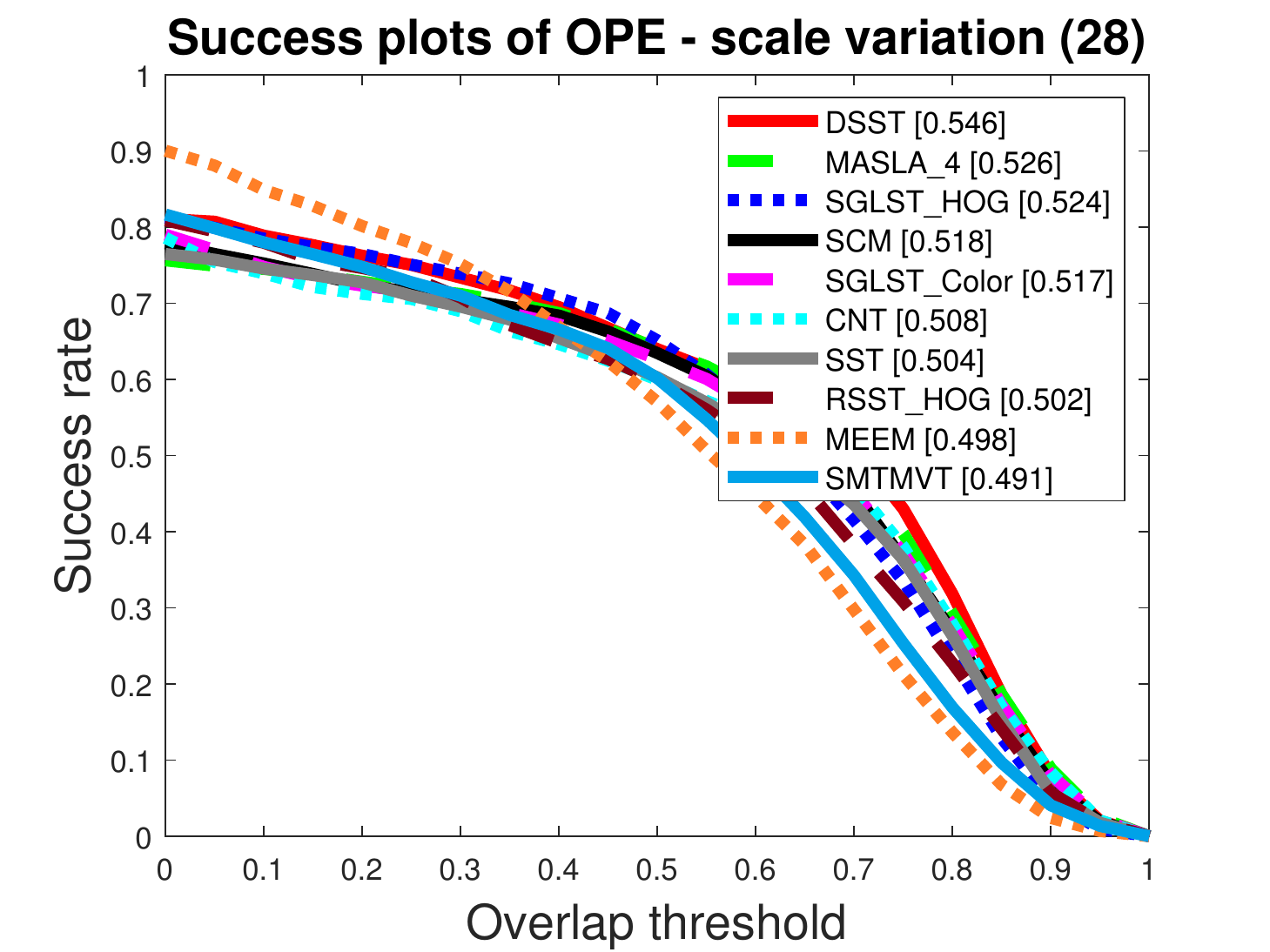}}
\caption{OTB50 OPE success plots for IV, LR, MB, OCC, OV, and SV challenge subsets. The value appearing in the title is the number of sequences in the specific subset. The values appearing in the legend are the AUC scores. Only the top 10 trackers are presented, while the results of the other trackers can be found in \cite{wu2013online}.}
\label{fig:fig3-2}
\end{figure*}
\begin{figure*}[t]
\centering     
\subfigure{\includegraphics[width=0.33\textwidth]{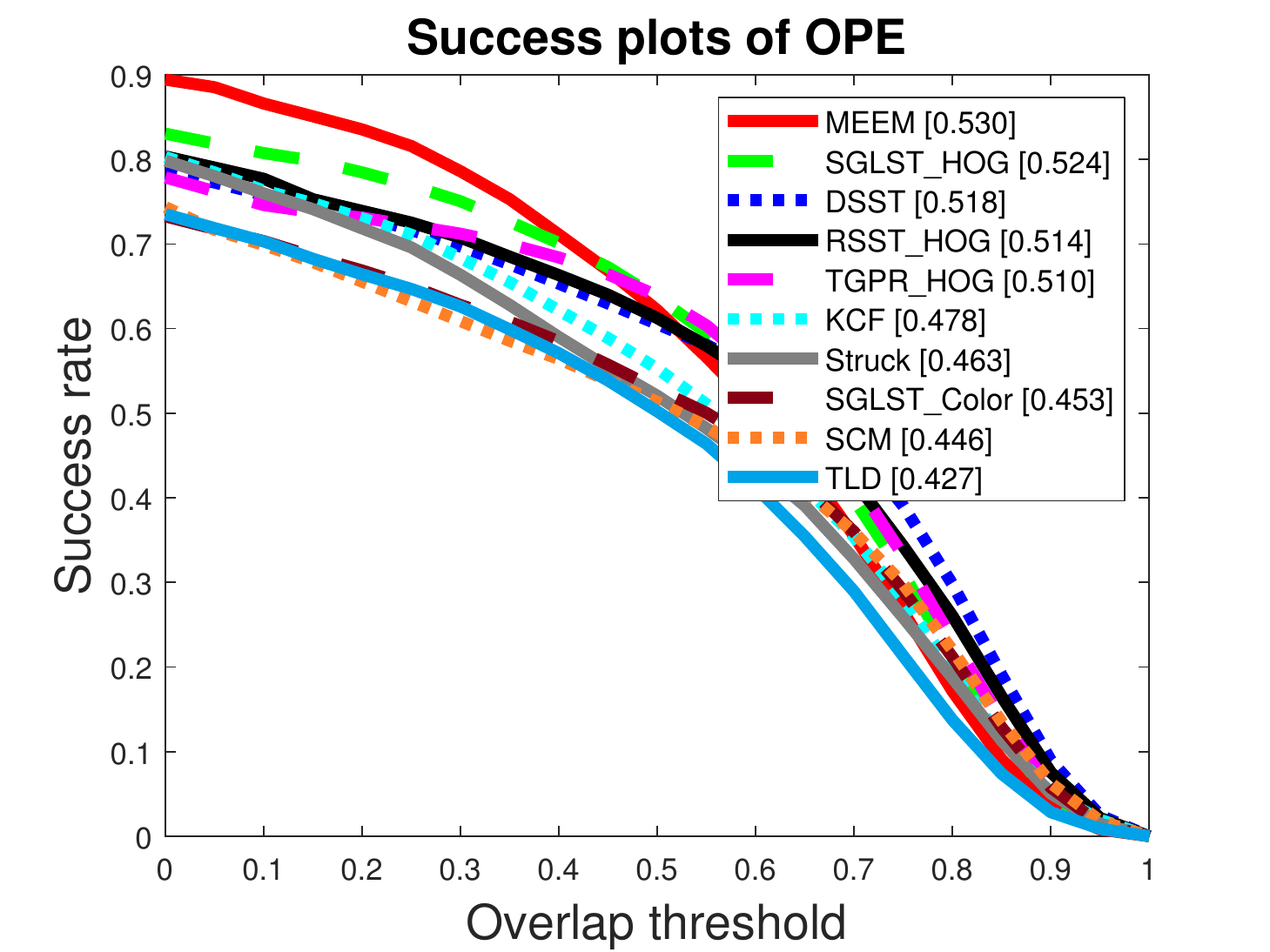}}
\subfigure{\includegraphics[width=0.33\textwidth]{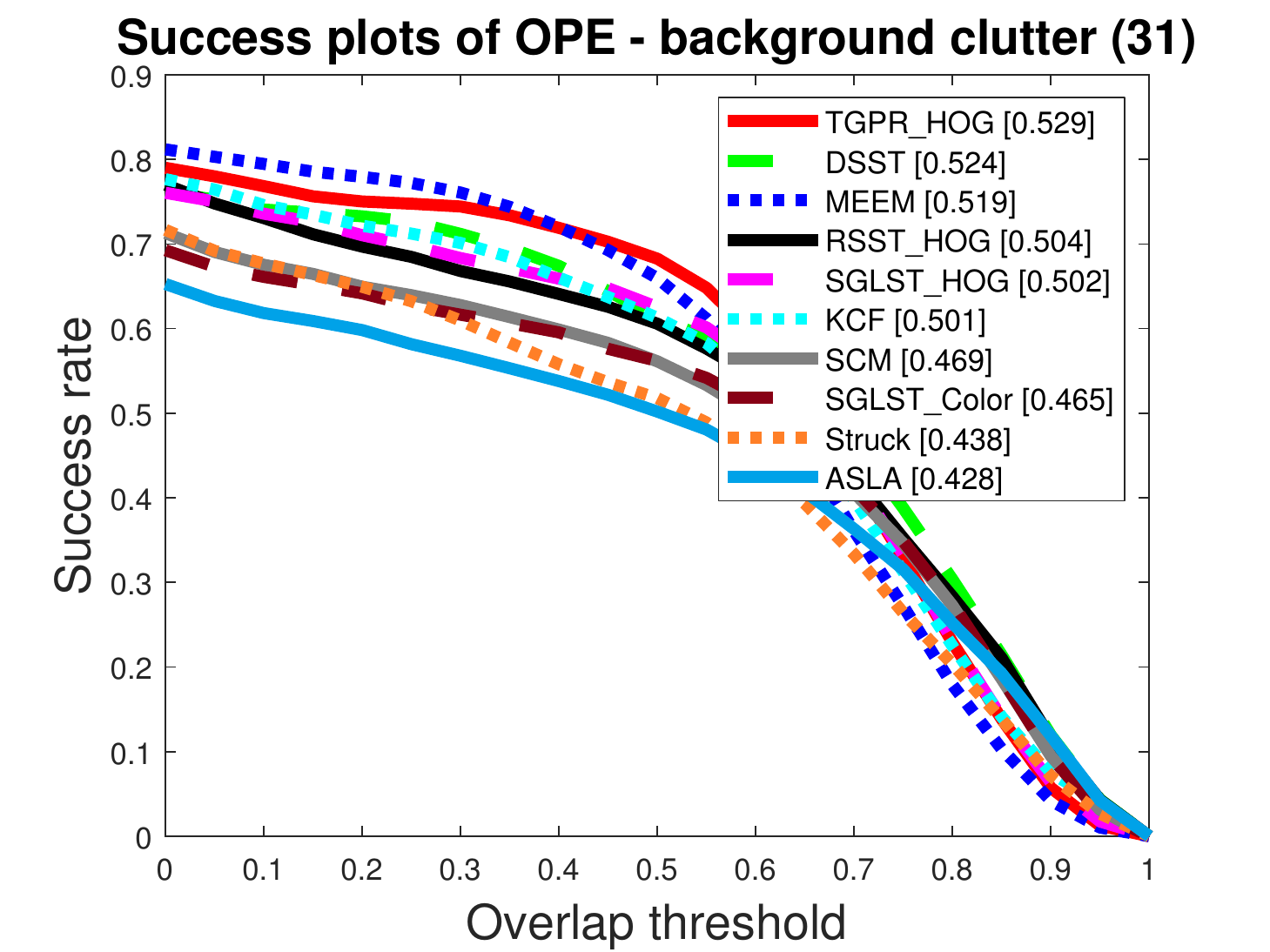}}
\subfigure{\includegraphics[width=0.33\textwidth]{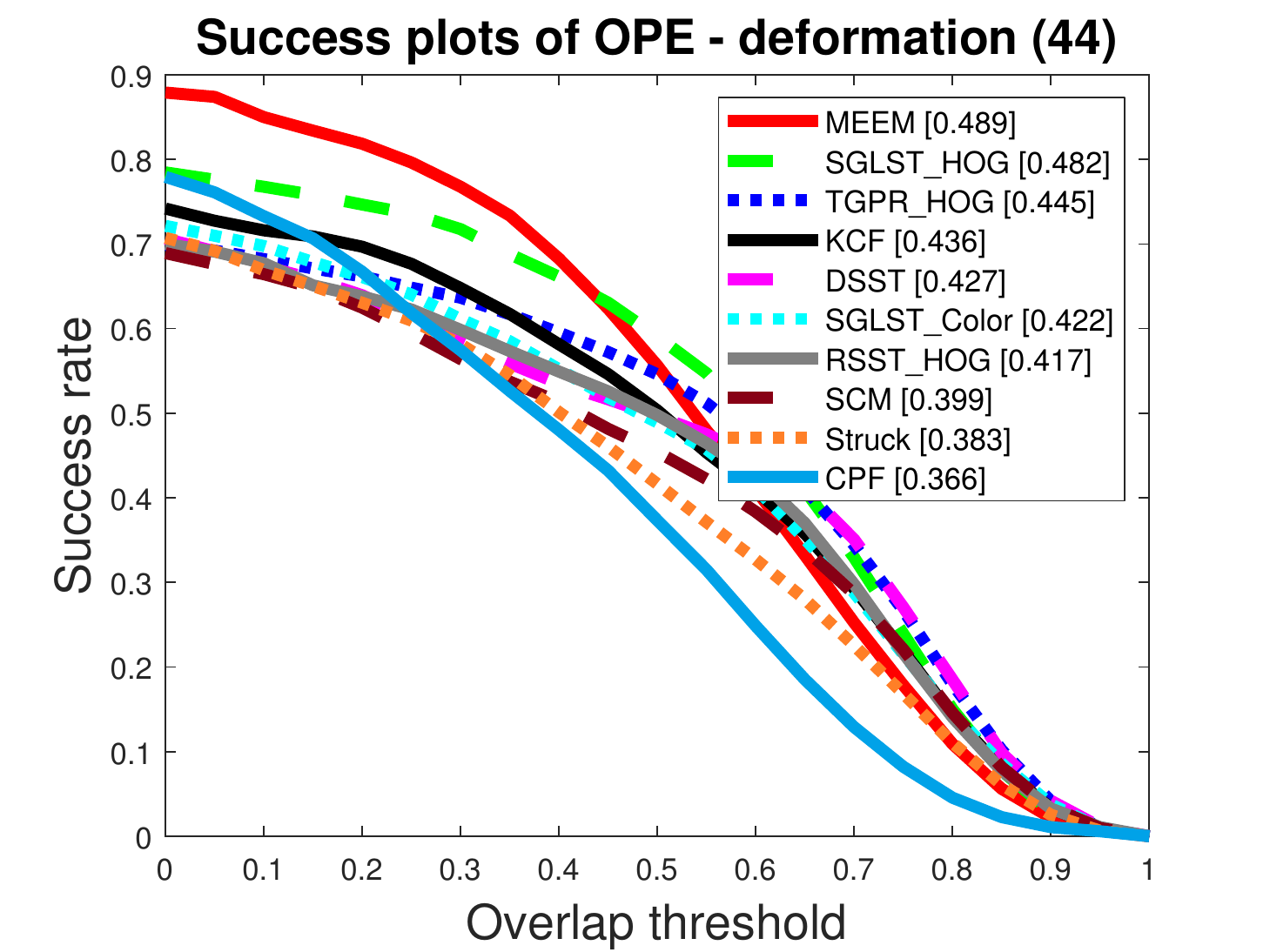}}
\subfigure{\includegraphics[width=0.33\textwidth]{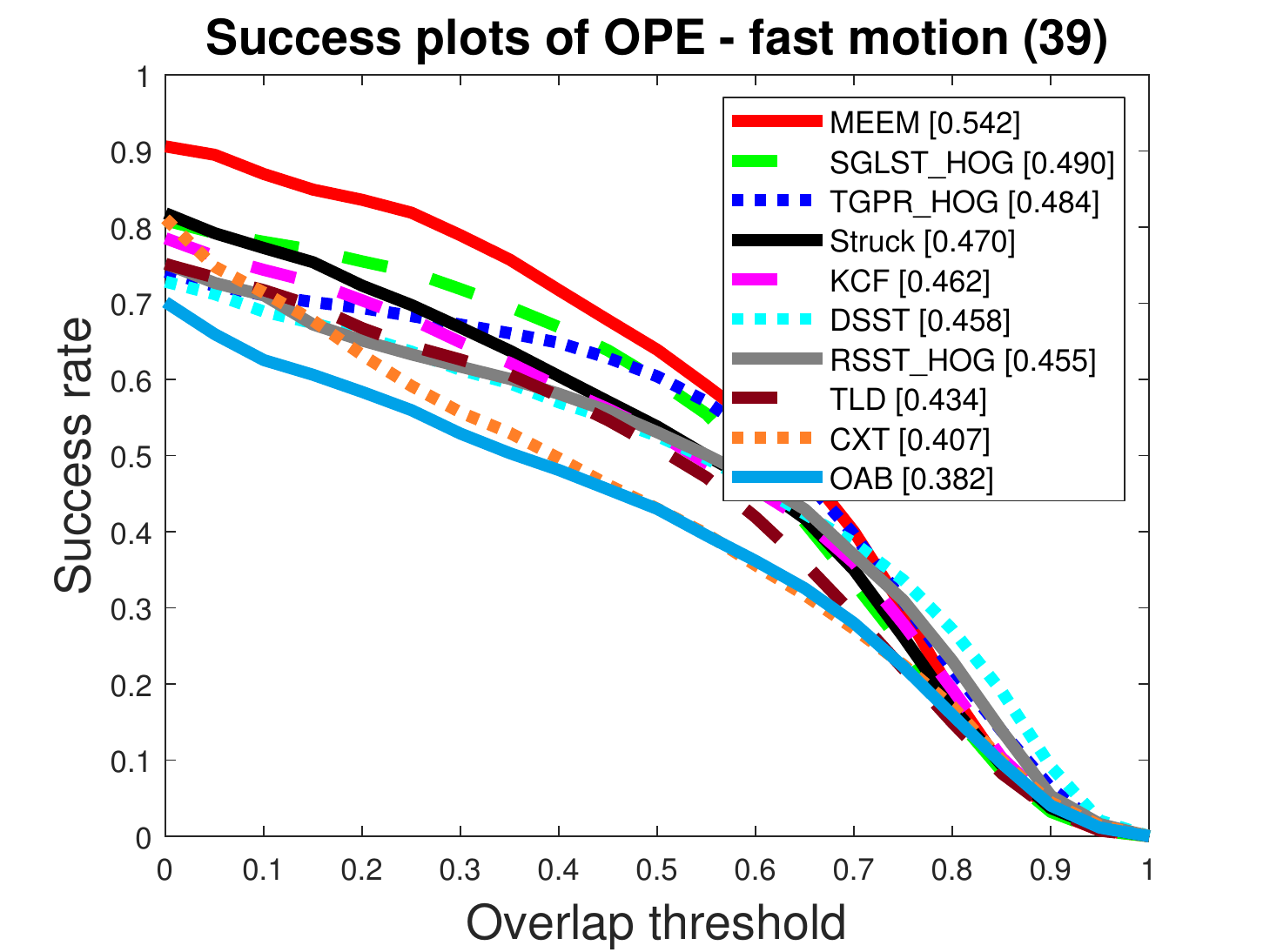}}
\subfigure{\includegraphics[width=0.33\textwidth]{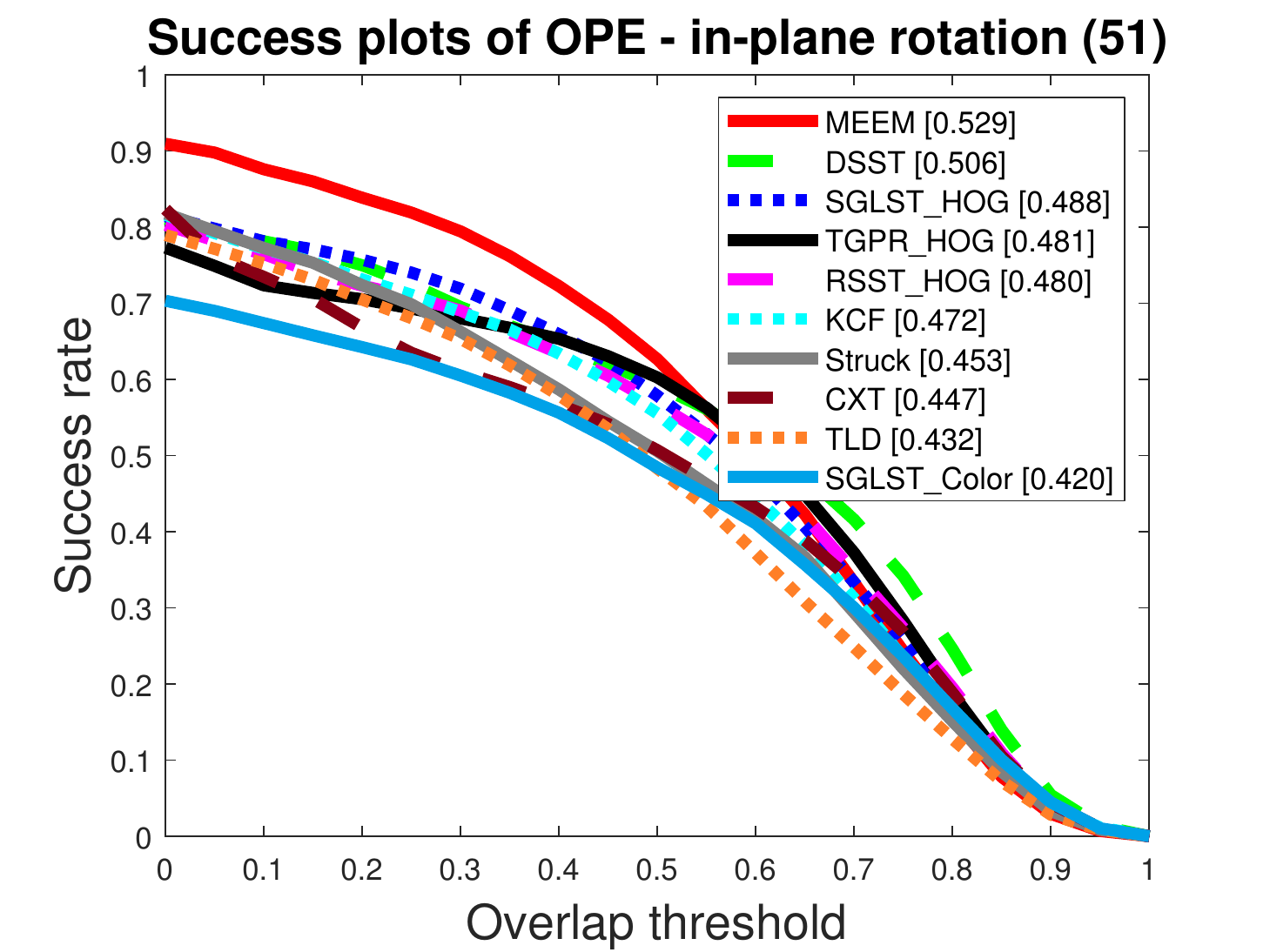}}
\subfigure{\includegraphics[width=0.33\textwidth]{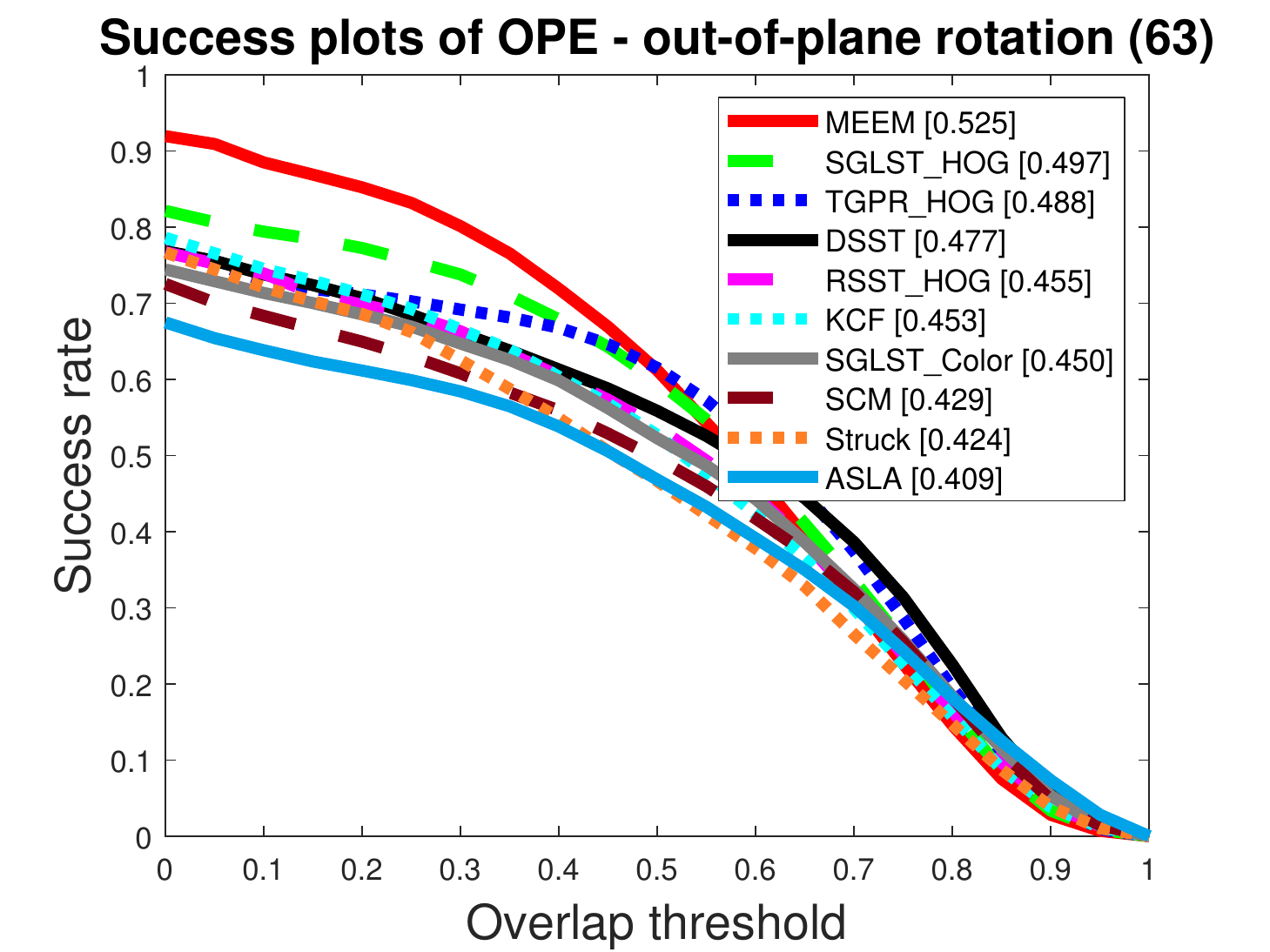}}
\caption{OTB100 overall OPE success plots and the OPE success plot BC, DEF, FM, IPR, and OPR challenge subsets. The value appearing in the title is the number of sequences in the specific subset. The values appearing in the legend are the AUC scores. Only the top 10 trackers are presented, while the results of the other trackers can be found in \cite{wu2015object}.}
\label{fig:fig4-1}
\end{figure*}
\subsection{Experimental Results on the OTB50 Benchmark}
We conduct the experiments on the OTB50 tracking benchmark \cite{wu2013online} to evaluate the overall performance of the proposed SGLST\_Color and SGLST\_HOG under different challenges. This benchmark consists of 50 annotated sequences, where 49 sequences has one annotated target and one sequence (\textit{jogging}) has two annotated targets. Each sequence is also labeled with attributes specifying the presence of different challenges including illumination variation (IV), scale variation (SV), occlusion (OCC), deformation (DEF), motion blur (MB), fast motion (FM), in-plane rotation (IPR), out-of-plane rotation (OPR), out-of-view (OV), background clutter (BC), and low resolution (LR). The sequences are categorized based on the attributes and 11 challenge subsets are generated. These subsets are utilized to evaluate the performance of trackers in different challenge categories. 

For this benchmark data set, there are online available tracking results for 29 trackers \cite{wu2013online}. In addition, we include the tracking results of additional 12 recent trackers, namely, MTMVTLS \cite{hong2013tracking}, MTMVTLAD \cite{mei2015robust}, MSLA-4 \cite{jia2016visual} (the recent version of ASLA \cite{jia2012visual}), SST \cite{zhang2015structural}, SMTMVT \cite{8486581}, CNT \cite{zhang2016robust}, TGPR \cite{gao2014transfer}, DSST \cite{danelljan2014accurate}, PCOM \cite{wang2014visual}, KCF \cite{henriques2015high}, MEEM\cite{zhang2014meem}, and RSST \cite{zhang2018robust}. Following the protocol proposed in \cite{wu2013online}, we use the same parameters for SGLST\_Color and SGLST\_HOG on all the sequences to 
obtain the one-pass evaluation (OPE) results, which are conventionally used to evaluate trackers by initializing them using the ground truth location in the first frame. We present the overall OPE success plot and the OPE success plots for BC, DEF, FM, IPR, and OPR challenge subsets in Figure \ref{fig:fig3-1} and the OPE success plots for IV, LR, MB, OCC, OV, and SV challenge subsets in Figure \ref{fig:fig3-2}. These success plots show the percentage of successful frames at the overlap thresholds ranging from 0 to 1, where the successful frames are the ones who have overlap scores larger than a given threshold. For fair comparison, we use the area under curve (AUC) of each success plot to rank the trackers. For convenience of the reader, we only include top 10 of the 43 compared trackers in each plot. The values in the parenthesis alongside the legends are AUC scores. The values in the parenthesis alongside the titles for 11 challenge subsets are the number of video sequences in the respective subset.
 It is clear from the overal success plot in Figure \ref{fig:fig3-1} that SGLST\_HOG (i.e., incorporating HOG features in SGLST) improves the tracking performance of SGLST\_Color (i.e., incorporating intensity features in SGLST). The similar improvement trends are also observed in \cite{henriques2012exploiting, zhang2018robust}. Among the 29 baseline trackers employed in \cite{wu2013online}, SCM achieves the most favorable performance. SGLST\_HOG outperforms SCM by 11.42\% in terms of the AUC score. Compared with the 12 additional recent trackers, SGLST\_HOG outperforms MSLA-4, SMTMVT, KCF, TGPR, RSST\_HOG, and CNT by 9.88\%, 9.66\%, 8.17\%, 5.10\%, 2.39\%, and 2.02\%, respectively. It achieves a comparable performance as that of DSST and MEEM. It should be mentioned that the variant of RSST with intensity features (i.e., RSST\_Color) reports the AUC score of 0.520 and the proposed SGLST\_Color achieves the AUC score of 0.523. This slight improvement indicates that the proposed optimization model is better than its counterpart in RSST\_Color.

The proposed SGLST\_HOG performs significantly better than traditional sparse trackers such as L1APG \cite{bao2012real}, LRST \cite{zhang2012low}, ASLA \cite{jia2012visual}, MTT \cite{zhang2012robust}, and MTMVTLS \cite{hong2013tracking}. It outperforms most recent sparse trackers such as MTMVTLAD \cite{mei2015robust}, SST \cite{zhang2015structural}, MSLA-4 \cite{jia2016visual}, SMTMVT \cite{8486581}, and RSST\_HOG \cite{zhang2018robust}. SGLST\_HOG, which yields the AUC score of 0.556, also achieves better performance than some correlation filter (CF) based methods such as KCF (AUC score of 0.514) and DSST (AUC score of 0.554). Moreover, it outperforms some deep learning-based methods such as CNT (AUC score of 0.545) and GOTURN (AUC score of 0.444) \cite{held2016learning}. However, the proposed SGLST\_HOG yields lower performance than some deep learning-based methods such as FCNT \cite{wang2015visual} (AUC score of 0.599), DLSSVM \cite{ning2016object} (AUC score of 0.589), and RSST\_Deep \cite{zhang2018robust} (AUC score of 0.590). We believe that SGLST can be further improved by incorporating the deep features, as the similar improvement trends are clearly shown in  RSST \cite{zhang2018robust}.

We further evaluate the performance of SGLST on 11 challenge subsets. As demonstrated in Figure \ref{fig:fig3-1} and Figure \ref{fig:fig3-2}, SGLST\_HOG ranks as one of the top three trackers in 5 subsets with DEF, OPR, LR, MB, and SV challenges and SGLST\_Color ranks as one of the top three trackers in 2 subsets with IV and LR challenges. SGLST\_HOG achieves the fourth rank on 2 subsets with IPR and OCC challenges and the fifth rank on 2 subsets with FM and IV challenges. SGLST\_Color achieves the fifth rank on one subset with the SV challenge. However, SGLST is not in the list of the top 10 trackers for the subset with the OV challenge. 
Overall, the proposed SGLST ranks as one of the top 5 trackers on 9 out of 11 subsets (e.g., 22 out of 50 image sequences) with DEF, OPR, LR, MB, SV, IV, IPR, OCC, and SV challenges. SGLST\_HOG significantly improves the tracking performance (i.e., the AUC score) of its variant and the third-ranked tracker, SGLST\_Color, by 23.31\% due to the incorporation of the HOG features instead of the intensity features.  It improves the tracking performance of the second-ranked tracker, RSST\_HOG, by 10.81\% mainly due to its novel optimization model that employs a group-sparsity regularization term to adopt local and spatial information
of the target candidates and attain the spatial layout structure among them.

\begin{figure*}[t]
\centering     
\subfigure{\includegraphics[width=0.33\textwidth]{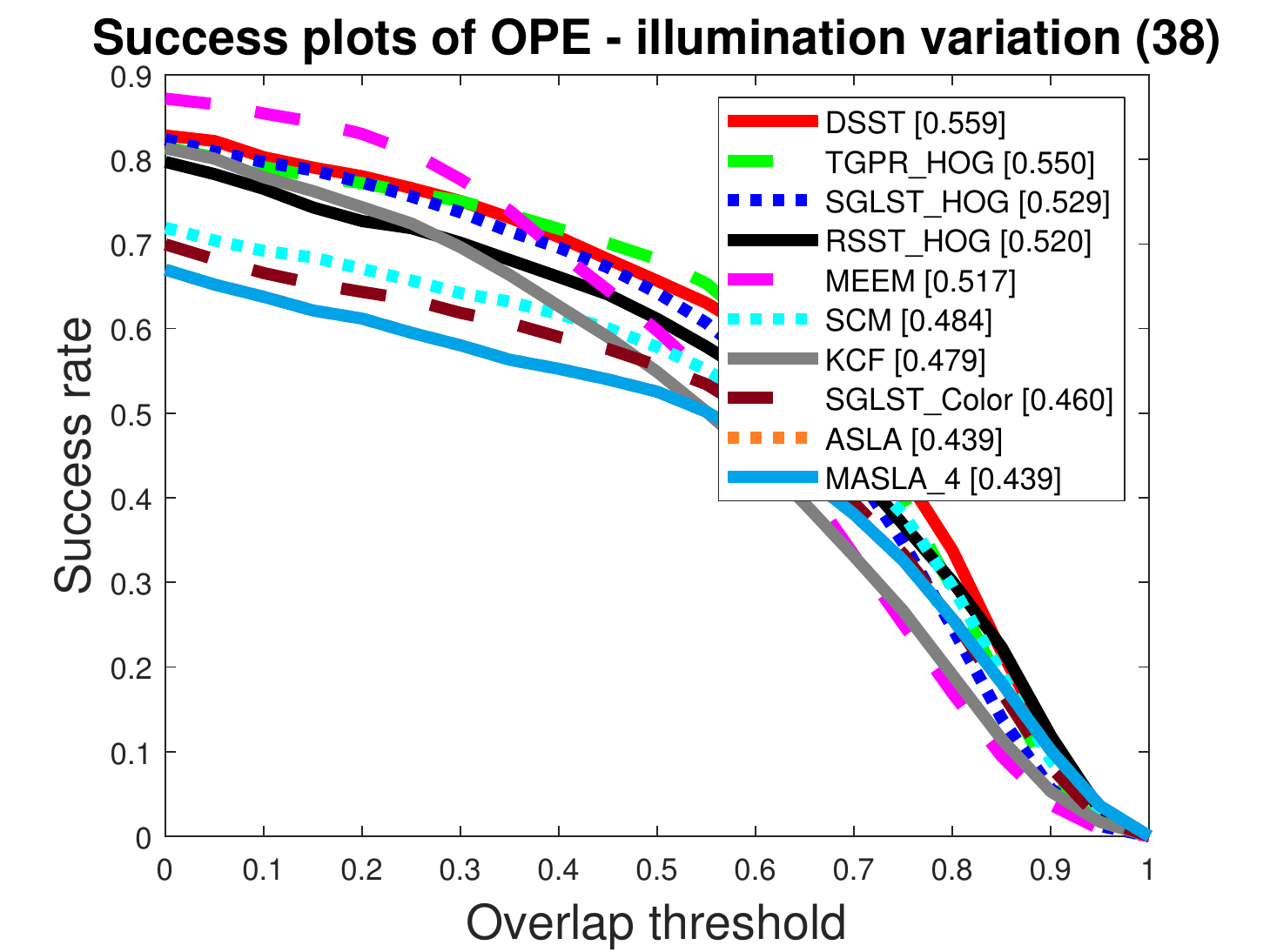}}
\subfigure{\includegraphics[width=0.33\textwidth]{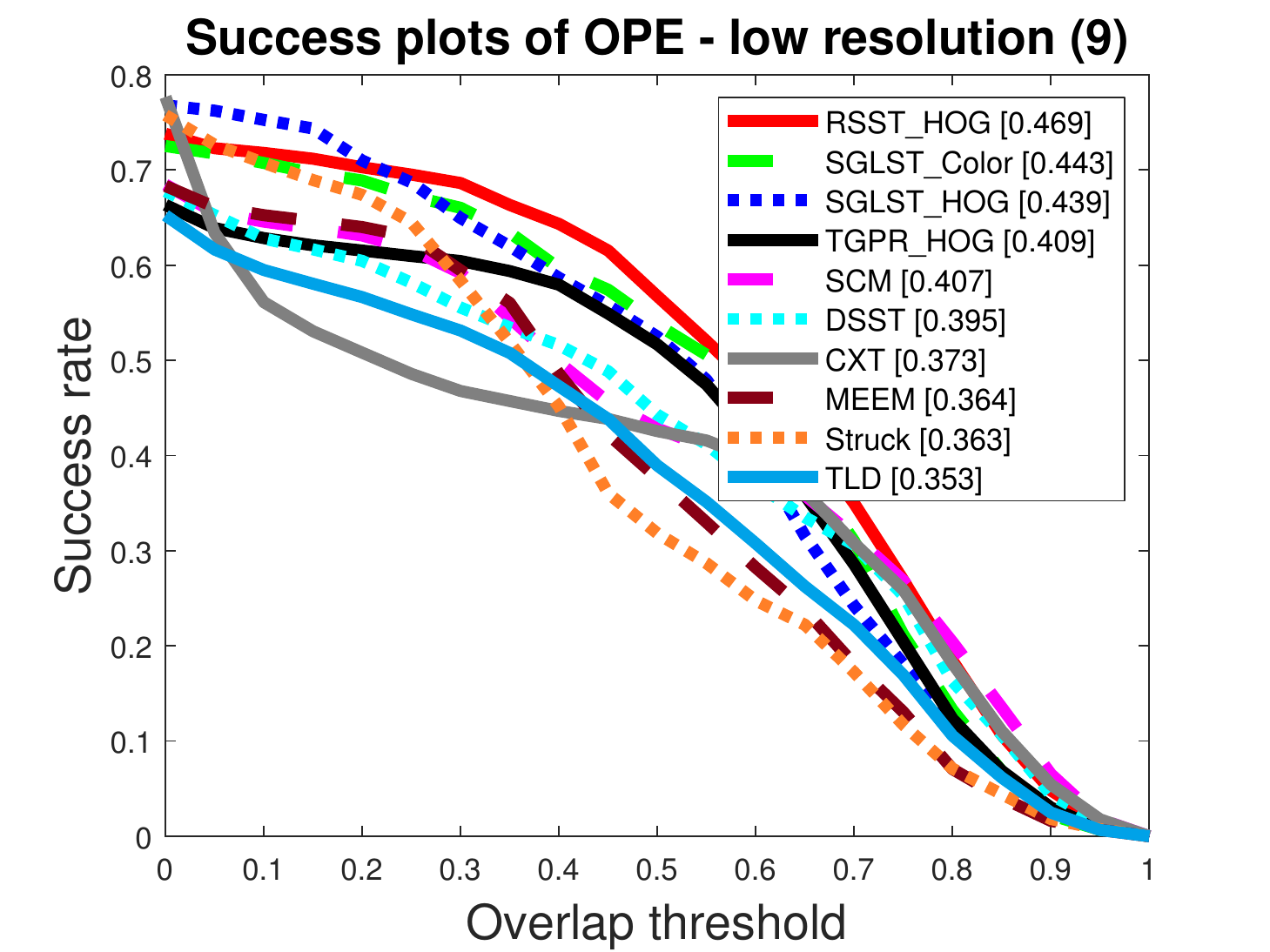}}
\subfigure{\includegraphics[width=0.33\textwidth]{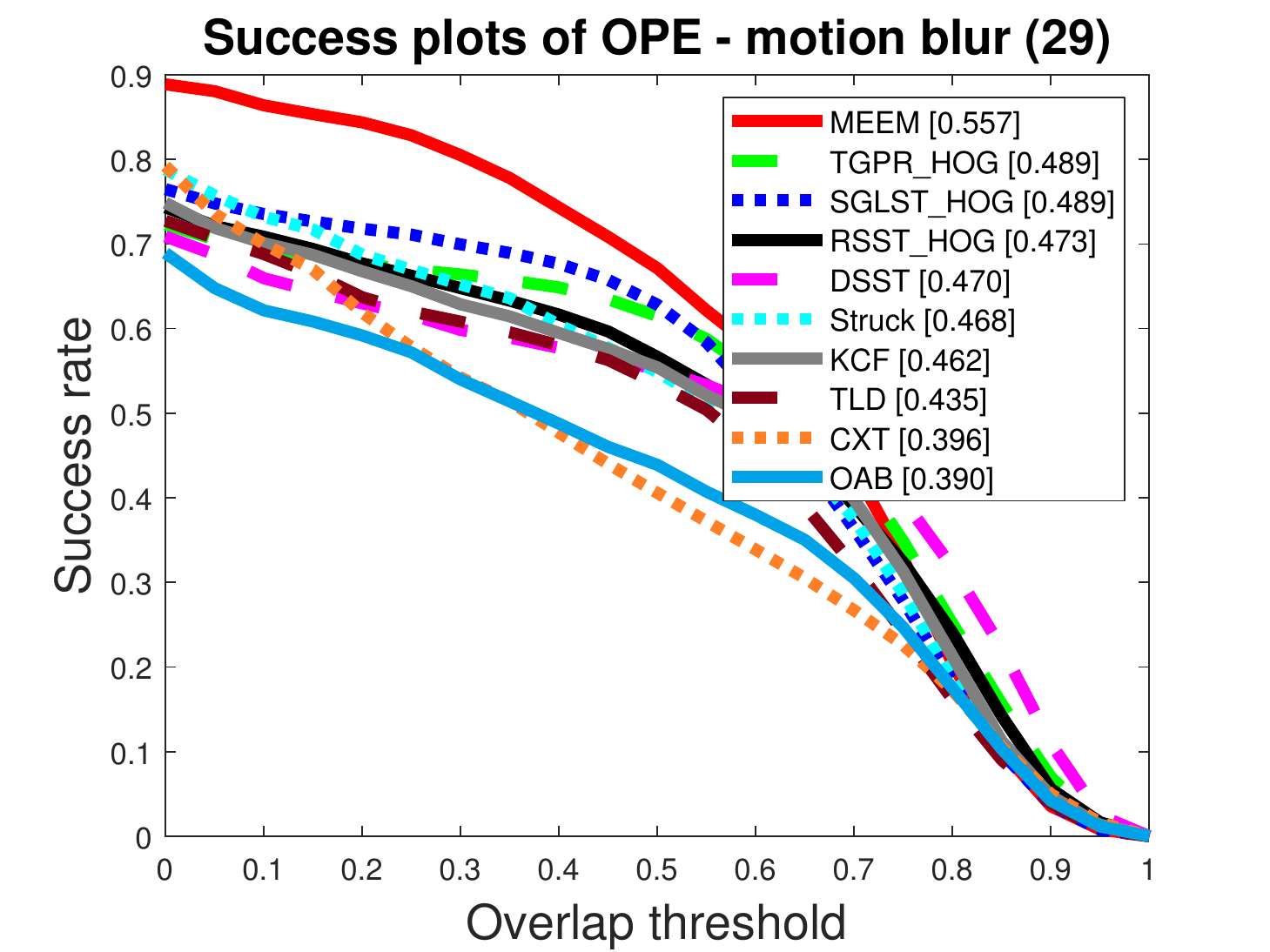}}
\subfigure{\includegraphics[width=0.33\textwidth]{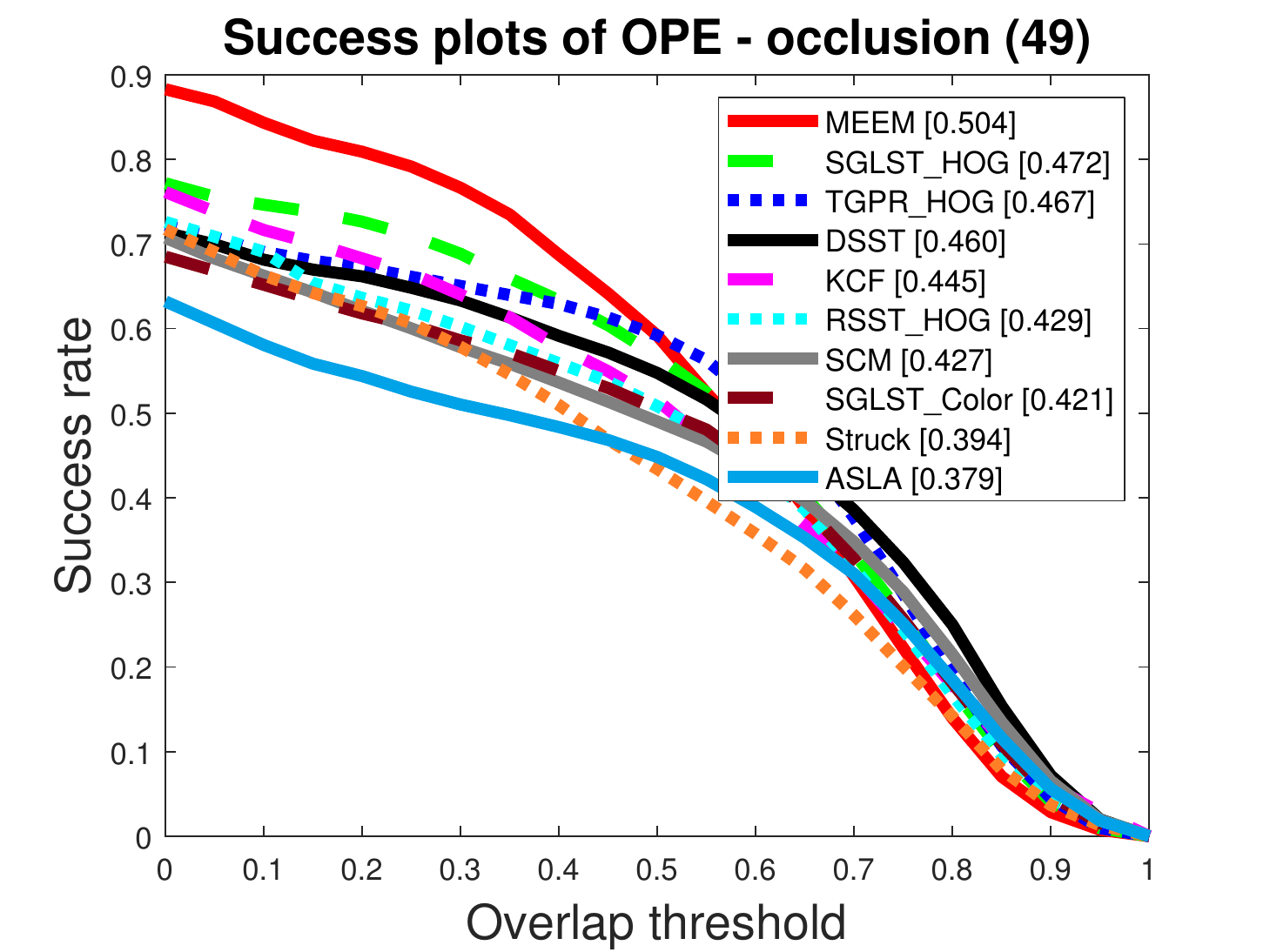}}
\subfigure{\includegraphics[width=0.33\textwidth]{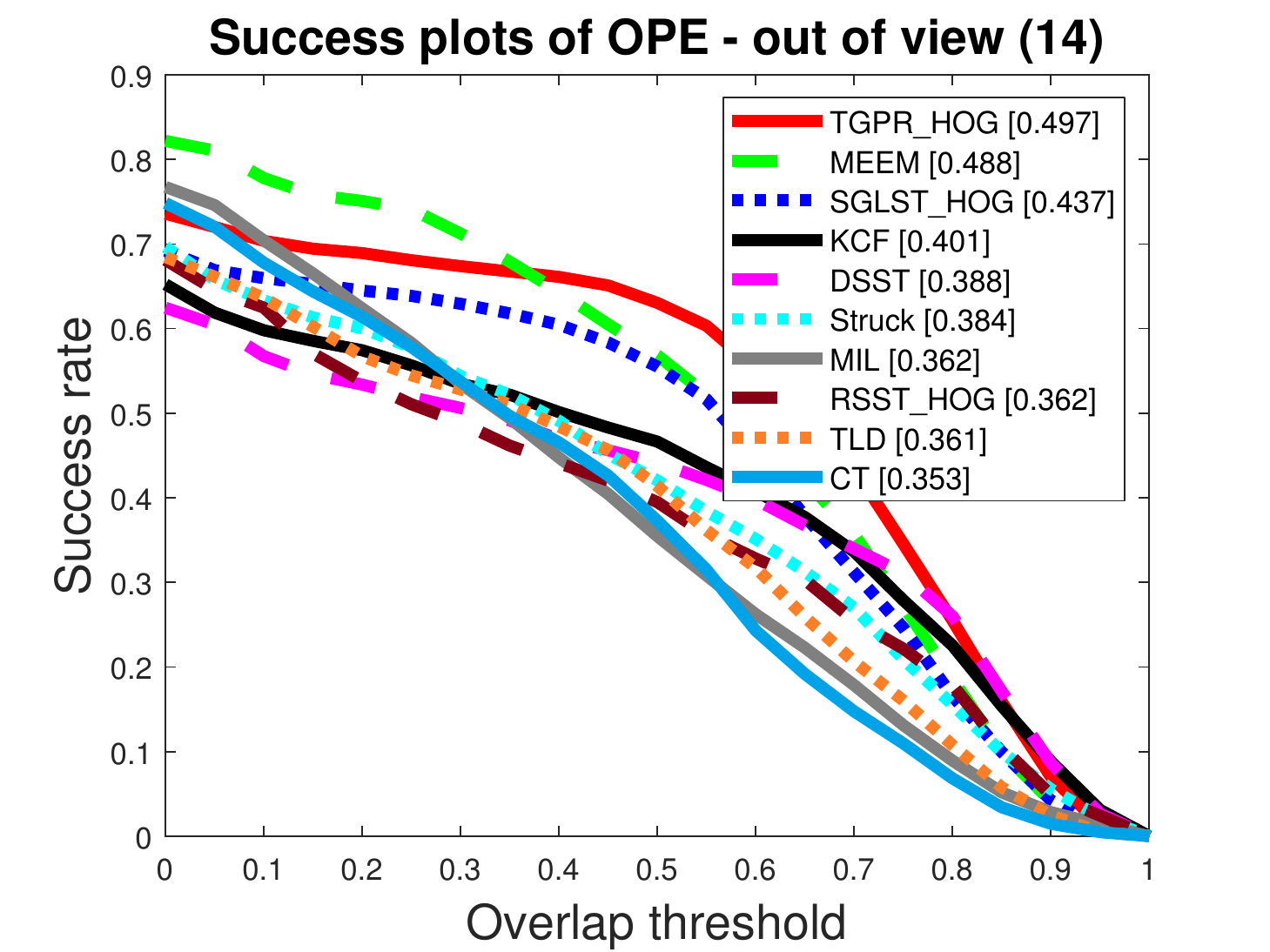}}
\subfigure{\includegraphics[width=0.33\textwidth]{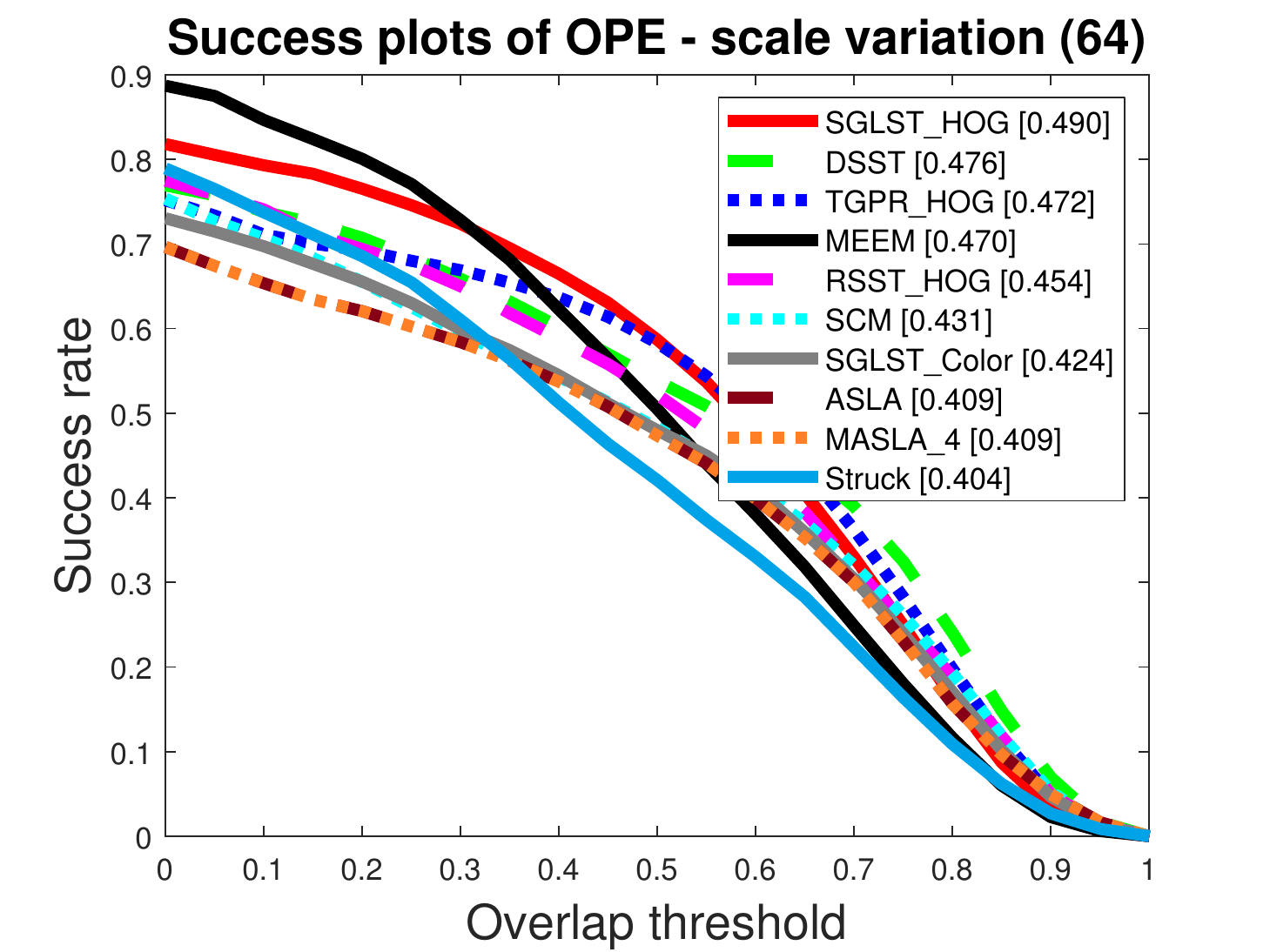}}
\caption{OTB100 OPE success plots for IV, LR, MB, OCC, OV, and SV challenge subsets. The value appearing in the title is the number of sequences in the specific subset. The values appearing in the legend are the AUC scores. Only the top 10 trackers are presented, while the results of the other trackers can be found in \cite{wu2015object}.}
\label{fig:fig4-2}
\end{figure*}
\subsection{Experimental Results on the OTB100 Benchmark}
We conduct the experiments on the OTB100 tracking benchmark \cite{wu2015object} to evaluate the overall performance of the proposed SGLST\_Color and SGLST\_HOG under different challenges. This benchmark is the extension of OTB50 \cite{wu2013online}, which consists of 100 annotated sequences. Each sequence is labeled with attributes specifying the presence of different challenges. The two sequences, \textit{jogging} and \textit{Skating}, have two annotated targets. The rest of 98 sequences have one annotated target. We evaluate the proposed SGLST against 29 baseline trackers used in \cite{wu2015object} and seven recent trackers including DSST \cite{danelljan2014accurate}, PCOM \cite{wang2014visual}, KCF \cite{henriques2015high}, MEEM \cite{zhang2014meem}, TGPR \cite{gao2014transfer}, and RSST \cite{zhang2018robust}.  The other 6 trackers compared in the OTB50 benchmark do not provide their results on the OTB100 benchmark.  Therefore, they are excluded in this experiment.

Figure \ref{fig:fig4-1} presents the overall OPE success plot and the OPE success plots for BC, DEF, FM, IPR, and OPR challenge subsets and Figure \ref{fig:fig4-2} provides the OPE success plots for IV, LR, MB, OCC, OV, and SV challenge subsets.  Top 10 trackers are included in each plot. The overall success plot in Figure \ref{fig:fig4-1} clearly demonstrates that the best tracker MEEM,  a multi-expert tracker employing an online linear SVM and an explicit feature mapping method, has a slightly better AUC score than the second best tracker, the proposed SGLST\_HOG. The difference in terms of the AUC score is only 0.006. SGLST\_HOG improves its variant, SGLST\_Color, by 16.67\% due to the use of HOG features over intensity features.  It also improves the fourth-ranked tracker RSST\_HOG, the most recent sparse tracker, by 1.95\% due to its novel optimization model. Compared to the third-ranked tracker DSST, a discriminative CF-based tracker, it improves the AUC score of DSST by 1.16\%. 

Similar to the tracking results obtained on the OTB50 tracking benchmark, the proposed SGLST\_HOG performs significantly better than traditional sparse trackers such as L1APG \cite{bao2012real}, LRST \cite{zhang2012low}, ASLA \cite{jia2012visual}, and MTT \cite{zhang2012robust}. It also outperforms RSST\_HOG \cite{zhang2018robust}, one of the most recent sparse trackers that provides the results on the OTB100 tracking benchmark. SGLST\_HOG, which yields the AUC score of 0.524, also achieves better performance than some CF and deep learning based methods such as KCF (AUC score of 0.478), DSST (AUC score of 0.518), and GOTURN (AUC score of 0.427) \cite{held2016learning}. However, it yields lower performance than some deep learning-based methods such as CNN-SVM (AUC of 0.554), CF2 (AUC of 0.562) \cite{ma2015hierarchical}, and RSST\_Deep (AUC of 0.583). We believe that incorporating the deep features in SGLST can further improve its tracking performance to be more comparable with the other deep-learning-based trackers.

We further evaluate the performance of SGLST on 11 challenge subsets in the OTB100 benchmark. As demonstrated in Figure \ref{fig:fig4-1} and Figure \ref{fig:fig4-2}, SGLST\_HOG ranks as one of the top three trackers in all 11 subsets except one subset with the BC challenge and SGLST\_Color ranks as one of the top three trackers in one subset with the LR challenge. SGLST\_HOG achieves the fifth rank on the subset with the BC challenges.  It achieves better performance than the best tracker, MEEM, in three subsets with IV, LR, and SV challenges.  Overall, the proposed SGLST ranks as the top 3 trackers on 10 out of 11 subsets (e.g., 69 out of 100 image sequences) with DEF, FM, IPR, OPR, IV, SV, LR, MB, OCC, and OV challenges.

\section{Conclusions}\label{sec:conc}
In this paper, we propose a novel tracker, called structured group local sparse tracker (SGLST), which exploits local patches within target candidates in the particle filter framework. Unlike conventional local sparse trackers, SGLST employs a new convex optimization model to preserve spatial layout structure among the local patches. To solve the proposed optimization model, we develop an efficient numerical algorithm consisting of two subproblems with closed-form solutions based on ADMM. We test the performance of the proposed tracker with two types of features including gray-level intensity features and HOG features. The qualitative and quantitative results on 16 publicly frame sequences demonstrate that SGLST\_HOG outperforms all compared state-of-the-art trackers. The experimental results on OTB50 and OTB100 tracking benchmarks demonstrate that SGLST\_HOG outperforms all compared state-of-the-art trackers except the MEEM tracker in terms of the average AUC score.
\bibliographystyle{IEEEbib}
\bibliography{egbib}
\end{document}